\documentclass[journal]{IEEEtran} 
\usepackage{times}
\usepackage{graphicx} 
\usepackage{subfigure} 
\usepackage{color}

\usepackage{authblk} 

\usepackage{algorithm}
\usepackage{algorithmic}

\usepackage{amsmath}
\usepackage{amssymb}

\newcommand{\norm}[1]{\left\Vert#1\right\Vert}
\newcommand{\abs}[1]{\left\vert#1\right\vert}
\newcommand\vect[1]{{\bf#1}}
\newcommand\matr[1]{{\bf#1}}
\newcommand{\Real}{\mathbb R}
\newcommand\RR[1]{\mathbb{R}^{#1}}
\newcommand\alphabf{{\boldsymbol{\alpha}}}
\newcommand\x{{\bf{x}}}
\newcommand\y{{\bf{y}}}
\newcommand\M{{\bf{M}}}
\newcommand\m{{\bf{m}}}
\newcommand\g{{\bf{g}}}
\newcommand\z{{\bf{z}}}
\newcommand\dist{{\psi}}

\newtheorem{thm}{Theorem}
\newtheorem{cor}[thm]{Corollary}

\newtheorem{prop}[thm]{Proposition}

\DeclareMathOperator{\sgn}{sgn}

\begin{document}

\title{Deep Neural Networks with Random Gaussian Weights: A Universal Classification Strategy?}

\author[*]{Raja Giryes}
\author[**]{Guillermo Sapiro}
\author[*]{Alex M. Bronstein}
\affil[*]{School of Electrical Engineering,
Faculty of Engineering,
Tel-Aviv University,
Ramat Aviv 69978, Israel. ~~~~~ 
{\tt\small \{raja@tauex, bron@eng\}.tau.ac.il}}
\affil[**]{Department of Electrical and Computer Engineering, 
Duke University, 
Durham, North Carolina, 27708, USA.
{\tt\small \{guillermo.sapiro\}@duke.edu}.}

\maketitle


\begin{abstract}
Three important properties of a classification machinery are: (i) the system preserves the core information of the input data; (ii) the training examples convey information about unseen data; 
and (iii) the system is able to treat differently points from different classes.
In this work we show that these fundamental properties are satisfied by the architecture of deep neural networks. We formally prove that these networks with random Gaussian weights perform a distance-preserving embedding of the data, with a special treatment for in-class and out-of-class data. Similar points at the input of the network are likely to have a similar output.
The theoretical analysis of deep networks here presented exploits tools used in the compressed sensing and dictionary learning literature, thereby making a formal connection between these important topics. The derived  results  allow drawing conclusions on the metric learning properties of the network and their relation to its structure, as well as providing bounds on
 the required  size of the training set such that the training examples would represent faithfully the unseen data. The results are validated with state-of-the-art trained networks.
\end{abstract}

\section{Introduction}

Deep neural networks (DNN) have led to a revolution in the areas of machine learning, audio analysis,  and computer vision, achieving state-of-the-art results in numerous applications \cite{Krizhevsky12ImageNet, Bengio13Representation,Schmidhuber15Deep}. 
In this work we formally study the properties of deep network
 architectures with random weights applied to data residing in a low dimensional manifold. 
 Our results provide insights into the outstanding empirically observed performance of DNN, the role of training, and the size of the training data.

Our motivation for studying networks with random weights is twofold. 
First, a series of works \cite{Pinto09High,Saxe11random, Beyond11Cox}
empirically showed successful DNN learning techniques based on randomization.
Second, studying a system with random weights rather than learned deterministic ones may lead to a better understanding of the system even in the deterministic case. For example, in the field of compressed sensing, where the goal is to recover a signal from a small number of its measurements, the study of random sampling operators led to breakthroughs in the understanding of the number of measurements required for achieving a stable reconstruction \cite{Candes06Near}. While the bounds provided in this case are universally optimal, the introduction of  a learning phase provides a better reconstruction performance as it adapts the system to the particular data at hand \cite{Elad07Optimized, Carvajalino09Learning, Hegde15NuMax}.
In the field of information retrieval, random projections have been used for locality-sensitive hashing (LSH) scheme capable of alleviating the curse of dimensionality for approximate nearest neighbor search in very high dimensions \cite{andoni2006near}. While the original randomized scheme is seldom used in practice due to the availability of data-specific metric learning algorithms, it has provided many fruitful insights.
Other fields such as phase retrieval, gained significantly from a study based on random Gaussian weights \cite{Candes13PhaseLift}.

Notice that the  technique of proving results for deep learning with assumptions on some random distribution and then showing that the same holds in the more general case is not unique to our work.  On the contrary, some of the stronger recent theoretical results on DNN follow this path.
For example, Arora \emph{et al.}  analyzed the learning of autoencoders with random weights in the range $[-1,1]$, showing that it is possible to learn them in polynomial time under some restrictions on the depth of the network \cite{Arora14Provable}.
Another example is the series of works \cite{Saxe14Exact, Choromanska15Loss, Dauphin14Identifying} that  study the optimization perspective of DNN.


In a similar fashion, in this work we study the properties of deep networks under the assumption of random weights. Before we turn to describe our contribution, we survey previous studies
 that formally analyzed the role of deep networks.

Hornik \emph{et al.} \cite{Hornik98Multilayer} and Cybenko \cite{Cybenko89Approximation} proved that neural networks serve as a universal approximation for any measurable Borel functions. However, finding the network weights for a given  function was shown to be
 NP-hard.

Bruna and Mallat proposed the wavelet scattering transform-- a cascade of wavelet transform convolutions with nonlinear modulus and averaging operators \cite{Bruna13Invariant}. They showed for this deep architecture that with more layers the resulting features can be made invariant to increasingly complex groups of transformations. 
The study of the wavelet scattering transform demonstrates that deeper architectures are able to better capture invariant properties of objects and scenes in images

Anselmi \emph{et al.}  showed that image representations invariant to transformations such as translations and scaling can considerably reduce the sample complexity of learning and that a deep architecture with filtering and pooling can learn such invariant representations \cite{Anselmi14Unsupervised}.
This result is particularly important in cases that training labels are scarce and in totally unsupervised learning regimes.

Mont{\'u}far and Morton showed that the depth of DNN allows representing restricted Boltzmann machines with a number of parameters exponentially greater than the number of the network parameters \cite{Montfar14When}. 
Mont{\'u}far \emph{et al.} suggest that each layer divides the space by a hyper-plane \cite{Montfar14Number}. Therefore, a deep network divides the space into an exponential number of sets, which is unachievable with a single layer with the same number of parameters.

Bruna \emph{et al.} showed that the pooling stage in DNN results in shift invariance \cite{Bruna13Learning}. In \cite{Bruna14Signal}, the same authors interpret this step as the removal of phase from a complex signal and show how the signal may be recovered after a pooling stage using phase retrieval methods.
This work also calculates the Lipschitz constants of the pooling and the rectified linear unit (ReLU) stages, showing that they perform a stable embedding of the data under the assumption that the filters applied in the network are frames, e.g., for the ReLU stage there exist two constants $0<A \le B$ such that for any $\x,\y \in \RR{n}$,
\begin{eqnarray}
A \norm{\x - \y}_2 \le \norm{\rho(\matr{M}\x) - \rho(\matr{M}\y)}_2 \le B \norm{\x - \y}_2,
\end{eqnarray} 
where $\matr{M} \in \RR{m \times n}$ denotes the linear operator at a given layer in the network  with $m$ and $n$ denoting the input and output dimensions, respectively, and $\rho(x) = \max(0, x)$ is the ReLU operator applied element-wise.
However, the values of the Lipschitz constants $A$ and $B$  in real networks and their behavior as a function of the data dimension currently elude understanding. To see why such a bound may be very loose, consider the output of only the linear part of a fully connected layer with random i.i.d. Gaussian weights, $m_{ij} \sim N(0,1/\sqrt{m})$, which is a standard initialization in deep learning. In this case, $A$ and $B$ scale like $\sqrt{\frac{2n}{m}}\left(1 \pm \sqrt{\frac{m}{2n}} \right)$  respectively
\cite{Rudelson10Non}. 
This undesired behavior is not unique  to a normally-distributed $\M$, being characteristic of any distribution with a bounded fourth  moment. Note that the addition of the non-linear operators, the ReLU and the pooling, makes these Lipschitz constants even worse.

\subsection{Contributions}

As the former example teaches, the scaling of the data introduced by $\M$ may  drastically deform the distances throughout each layer, even in the case where $m$ is very close to $n$, which makes it unclear whether it is possible to recover the input of the network (or of each layer) from its output. 
In this work, the main question we focus on is: What happens to the metric of the input data throughout the network? We focus on the above mentioned setting, assuming that the network has random i.i.d. Gaussian weights. 
We prove that DNN preserve the metric structure of the data as it propagates along the layers, allowing for the stable recovery of
 the original data from the features calculated by the network. 

This type of property is often encountered in the literature \cite{Bruna14Signal, Mahendran15Understanding}. Notice, however, that the recovery of  the input is possible if the size of the network output is proportional to the intrinsic dimension of the data at the input (which is not the case at the very last layer of the network, where we have class labels only), similarly to data reconstruction from a small number of random projections \cite{Chandrasekaran12Convex, Amelunxen14Living, Plan13Robust}.
However, unlike random projections that preserve the Euclidean distances up to a small distortion \cite{Johnson84Extensions}, each layer of DNN with random weights distorts these distances proportionally to the angles between  its input points: the smaller the angle at the input, the stronger the shrinkage of the distances. Therefore, the deeper the network, the stronger the shrinkage we get. 
Note that this does not contradict the fact that we can recover the input from the output; even when properties such as lighting, pose and location are removed from an image (up to certain extent), the resemblance to the original image is still maintained.

As random projection is a universal sampling strategy for any low dimensional data \cite{Candes06Near,Plan13Robust,Mendelson08Uniform}, deep networks with random weights are a universal system that separates any data (belonging to a low dimensional model) according to the angles between its points, where the general assumption is that there are large angles between different classes \cite{Yamaguchi98Face, Wolf03Learning, Elhamifar13Sparse, Qiu15Learning}. As training of the projection matrix adapts it to better preserve specific distances over others, the training of a network prioritizes intra-class angles over inter-class ones. This relation is alluded by our proof techniques and is empirically manifested by observing the angles and the Euclidean distances at the output of trained networks, as demonstrated later in the paper in Section~\ref{sec:training}.

The rest of the paper is organized as follows: In Section~\ref{sec:model}, we start by utilizing the recent theory of 1-bit compressed sensing to show that each DNN layer preserves the metric of its input data in the Gromov-Hausdorff sense up to a small constant $\delta$, under the assumption that these data reside in a low-dimensional manifold denoted by $K$. This allows us to draw conclusions on the tessellation of the space created by each layer of the network and the relation between the operation of these layers and local sensitive hashing (LSH) \cite{andoni2006near}. 
We also show that it is possible to retrieve the input of a layer, up to certain accuracy, from its output. This implies that every layer preserves the important information of the data.

In Section~\ref{sec:dist_ang}, we proceed by  analyzing the behavior of the Euclidean distances and angles in the data throughout the network.
This section reveals an important effect of the ReLU. Without the ReLU, we would just have random projections and Euclidean distance preservation. Our theory shows that the addition of ReLU makes the system sensitive to the angles between points.
We prove that networks tend to decrease the Euclidean distances between points with a small angle between them (``same class''), more than the distances between points with large angles between them (``different classes'').

Then, in Section~\ref{sec:ent_net} we prove that low-dimensional data at the input remain such throughout the entire network, i.e., DNN (almost) do not increase the intrinsic dimension of the data. This property is used in Section~\ref{sec:meas_num} to deduce the size of data needed for training DNN. 

We conclude by studying the role of training in Section~\ref{sec:training}. As random networks are blind to the data labels, training may select discriminatively the angles that cause the distance deformation. Therefore, it will cause distances between different classes to increase more than the distances within the same class. We demonstrate this in several simulations, some of which with networks that recently showed state-of-the-art performance on challenging datasets, e.g., the network by \cite{Simonyan15VeryDeep} for the ImageNet dataset \cite{ImageNet14}.  Section~\ref{sec:conc} concludes the paper by summarizing the main results and outlining future research directions.

It is worthwhile emphasizing that the assumption that classes are separated by large angles is common in the literature (see \cite{Yamaguchi98Face, Wolf03Learning, Elhamifar13Sparse, Qiu15Learning}). This assumption can further refer to some feature space rather than to the raw data space. Of course, some examples might be found that contradict this assumption such as the one of two concentric spheres, where each sphere represents a different class. With respect to such particular examples two things should be said: First, these cases are rare in real life signals, typically exhibiting some amount of scale invariance that is absent in this example. Second, we prove that the property of discrimination based on angles holds for DNN with random weights and conjecture in Section~\ref{sec:training} that a potential role (or consequence) of training in DNN is to favor certain angles over others and to select the origin of the coordinate system with respect to which the angles are calculated. We illustrate in Section~\ref{sec:training} the effect of training compared to random networks, using trained DNN that have achieved state-of-the-art results in the literature. Our claim is that DNN are suitable for models with clearly distinguishable angles between the classes if random weights are used, and for classes with some distinguishable angles between them if training is used.

For the sake of simplicity of the discussion and presentation clarity, 
we focus only on the role of the ReLU operator \cite{Nair10Rectified}, assuming that our data are properly aligned, i.e., they are not invariant to operations such as rotation or translation, and therefore there is no need for the pooling operation to achieve invariance.
Combining recent results for the phase retrieval problem with the proof techniques in this paper can lead to a theory also applicable to the pooling operation. In addition, with the strategy in  \cite{Haupt10Toeplitz, Aperiodic12Saligrama, Rauhut12Restricted, Ai14One}, it is possible to generalize our guarantees to sub-Gaussian distributions  and to random convolutional filters.  We defer these natural extensions to future studies.

\section{Stable Embedding of a Single Layer}
\label{sec:model}

In this section we consider the distance between the metrics of the input and output of a single DNN layer of the form $f(\matr{M} \cdot )$,
mapping an input vector $\x$ to the output vector $f(\M\x)$,
 where $\M$ is an $m\times n$ random Gaussian matrix and $f : \Real \rightarrow \Real $ is a semi-truncated linear function applied element-wise. $f$ is such if it is linear on some (possibly, semi-infinite) interval and constant outside of it, $f(0) = 0$, $0 < f(x) \le x, \forall x > 0$, and $0 \ge f(x) \ge x, \forall x < 0$. The ReLU, henceforth denoted by $\rho$,
 is an example of such a function, while the sigmoid and the hyperbolic tangent functions satisfy this property approximately. 

\begin{figure}[ht]
\begin{center}
{
{\includegraphics[width=0.33\textwidth]{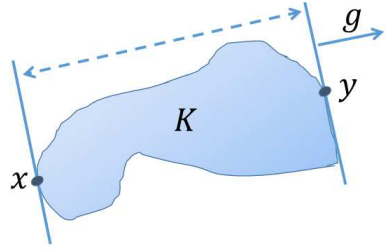}}}
\end{center}
\caption{Width of the set $K$ in the direction of $\vect{g}$. This figure is a variant of Fig.~1 in \cite{Plan13Robust}. Used with permission of the authors.}
\label{fig:GaussianWidth}
\end{figure}

We assume that the input data belong to a manifold $K$ with the Gaussian mean width
\begin{eqnarray}
\omega(K) := \mathbb{E}[ \sup_{\vect{x}, \vect{y} \in K} \langle \vect{g}, \vect{x} - \vect{y} \rangle ],
\end{eqnarray}
where the expectation is taken over a random vector $\vect{g}$ with normal i.i.d. elements. 
To better understand this definition, note that $\sup_{\vect{x}, \vect{y} \in K} \langle \vect{g}, \vect{x} - \vect{y} \rangle$ is the width of the set $K$ in the direction of $\vect{g}$ as illustrated in Fig.~\ref{fig:GaussianWidth}. The mean provides an average over the widths of the set $K$ in different isotropically distributed directions, leading to the definition of the Gaussian mean width $\omega(K)$.

The Gaussian mean width is a useful measure for the dimensionality of a set. As an example we consider the following two popular data models:

\paragraph{Gaussian mixture models} $K \subset \mathbb{B}^2$ (the $\ell_2$-ball) consists of $L$ Gaussians of dimension $k$. In this case $\omega(K) =  O(\sqrt{k + \log{L}})$.

\paragraph{Sparsely representable signals}
The data in $K \subset \mathbb{B}^2$ can be approximated by a sparse linear combination of atoms of a dictionary, i.e., 
$K = \left\{ \x = \matr{D}\alphabf : \norm{\alphabf}_0 \le k, \norm{\x}_2 \le 1
\right\}$, where $\norm{\cdot}_0$ is the pseudo-norm that counts the number of non-zeros in a vector and $\matr{D}\in \RR{n \times L}$ is the given dictionary. For this model $\omega(K) = O(\sqrt{k\log(L/k)})$.

Similar results can be shown for other models such as union of subspaces
and low dimensional manifolds. For more examples and details on $\omega(K)$, we refer the reader to \cite{Plan13Robust, Plan14Dimension}.

We now show that each standard DNN layer performs 
a stable embedding of the data in the Gromov-Hausdorff sense \cite{Gromov99Metric}, i.e., it is a $\delta$-isometry between $(K,d_K)$ and $(K',d_{K'})$, where $K$ and $K'$ are the manifolds of the input and output data and $d_K$ and $d_{K'}$ are metrics induced on them. A function $h:K\rightarrow K'$ is a $\delta$-isometry  if 
\begin{eqnarray}
\abs{d_{K'}(h(\x), h(\y) )  - d_K(\x, \y )} \le \delta, \forall \x,\y \in K,
\end{eqnarray}
and for every $\x' \in K'$ there exists $\x \in K$ such that $d_Y(\x',h(\x)) \le \delta$ (the latter property is sometimes called $\delta$-surjectivity).
In the following theorem and throughout the paper $C$ denotes a given constant (not necessarily the same one) and $\mathbb{S}^{n-1}$ the unit sphere in $\RR{n}$. 

%

\begin{thm}
\label{thm:deep_net_stable}
Let $\matr{M}$ be the linear operator applied at a DNN layer, $f$ a semi-truncated linear activation function, and $K \subset \mathbb{S}^{n-1}$ the manifold of the input data for that layer.  If $\sqrt{m}\matr{M} \in \RR{m \times n}$ is a random matrix with i.i.d normally distributed entries,
then the map $g : (K \subset \mathbb{S}^{n-1},d_{\mathbb{\,S}^{n-1}}) \rightarrow (g(K), d_{\,\mathbb{H}^m})$ defined by
$g(\x) = \sgn(f(\M\vect{x}))$
is with high probability a $\delta$-isometry, i.e., 
$$
\left| d_{\mathbb{\,S}^{n-1}}(\x,\y) - d_{\,\mathbb{H}^m}( g(\x), g(\y) )  \right| \le \delta, ~ \forall \x,\y \in K,  
$$
with $\delta \le C\, m^{-1/6} \,\omega^{1/3} (K)$.
\end{thm}

In the theorem $d_{\mathbb{S}^{n-1}}$ is the geodesic distance on $\mathbb{S}^{n-1}$, $d_{\mathbb{H}}$ is the Hamming distance and the sign function, $\sgn(\cdot)$, is applied elementwise, and is defined as $\sgn(x) = 1$ if  $x>0$ and $\sgn(x) = -1$ if $x\le 0$.
The proof of the approximate injectivity follows from the proof of Theorem~1.5 in \cite{Plan14Dimension}.

Theorem~\ref{thm:deep_net_stable} is important as it provides a better understanding of the tessellation of the space that each layer creates.
This result stands in line with \cite{Montfar14Number} that suggested that each layer in the network creates a tessellation of the input data by the different hyper-planes imposed by the rows in $\M$. 
However, Theorem~\ref{thm:deep_net_stable} implies more than that. It implies that each cell in the tessellation has a diameter of at most $\delta$ (see also Corollary~1.9 in \cite{Plan14Dimension}), i.e., if $\x$ and $\y$ fall to the same side of all the hyperplanes, then $\norm{\x - \y}_2 \le \delta$. In addition, the number of hyperplanes separating two points $\x$ and $\y$ in $K$ contains enough information to deduce their distance up to a small distortion.
From this perspective, each layer followed by the sign function acts as locality-sensitive hashing \cite{andoni2006near},  approximately embedding the input metric into the Hamming metric. 

Having a stable embedding of the data, it is natural to assume that it is possible to recover the input of a layer from its output. Indeed, Mahendran and Vedaldi demonstrate that it is achievable through the whole network \cite{Mahendran15Understanding}. 
The next result provides a theoretical justification for this, showing that it is possible to recover the input of each layer from its output:

\begin{thm}
\label{thm:deep_net_stable_rec}
Under the assumptions of Theorem~\ref{thm:deep_net_stable} there exists a program $\mathcal{A}$ such that 
$\norm{\vect{x} - \mathcal{A}(f(\matr{M}\vect{x} ) )}_2 \le \epsilon$, 
where $\epsilon = O\left(\frac{\omega(K)}{\sqrt{m}}\right)$.
\end{thm}

The proof follows from Theorem~1.3 in \cite{Plan13Robust}.
If $K$ is a cone then one may use also Theorem~2.1 in \cite{Plan14High} to get a similar result.

Both theorems~\ref{thm:deep_net_stable} and \ref{thm:deep_net_stable_rec} are applying existing results from 1-bit compressed sensing on DNN. Theorem~\ref{thm:deep_net_stable} deals with embedding into the Hamming cube and Theorem~\ref{thm:deep_net_stable_rec} uses this fact to show that we can recover the input from the output. Indeed, Theorem~\ref{thm:deep_net_stable} only applies to an individual layer and cannot be applied consecutively throughout the network,  since it deals with embedding into the Hamming cube. One way to deal with this problem is to extend the proof in \cite{Plan14Dimension} to an embedding from $\mathbb{S}^{n-1}$ to $\mathbb{S}^{m-1}$. Instead, we turn to focus on the ReLU and prove more specific results about the exact distortion of angles and Euclidean distances. These also include a proof about a stable embedding of the network at each layer from $\RR{n}$ to $\RR{m}$.

\section{Distance and Angle Distortion}
\label{sec:dist_ang}

So far we have focused on the metric preservation of the deep networks in terms of the Gromov-Hausdorff distance. 
In this section we turn to look at how the Euclidean distances and angles change within the layers. We focus on the case of ReLU as the activation function.
A similar analysis can also be applied for pooling.  For the simplicity of the discussion we defer it to future study. 

\begin{figure}[t]
\begin{center}
{
{\includegraphics[width=0.3\textwidth]{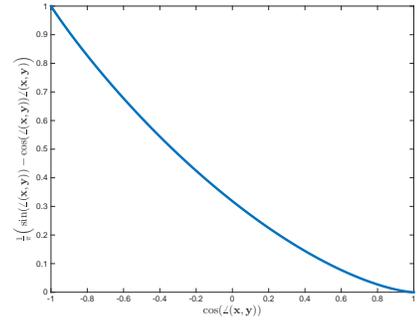}}}
\end{center}
\caption{Behavior of $\dist(\x,\y)$  (Eq. \eqref{eq:euclead_add_term}) as a function of $\cos \angle (\x, \y)$. The two extremities $\cos \angle (\x, \y) = \pm 1$  correspond to the angles zero  and $\pi$, respectively,  between $\x$ and $\y$. Notice that the larger is the angle between the two, the larger is the value of $\dist(\x,\y)$. In addition, it vanishes for small angles between the two vectors.}
\label{fig:cos_sin_term_behav}
\end{figure}

\begin{figure}[htb]
\begin{center}
{
{\includegraphics[width=0.3\textwidth]{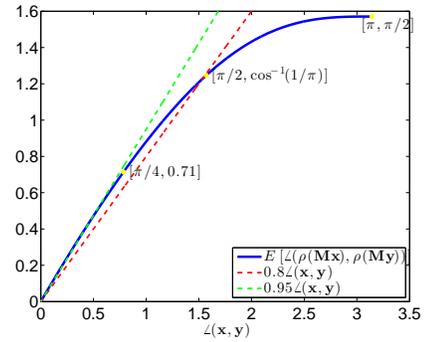}}}
\end{center}
\caption{Behavior of the angle between two points $\x$ and $\y$  in the output of a DNN layer as a function of their angle in the input. In the ranges  $[0,\pi/4]$ and $[\pi/4, \pi/2]$ the output behaves approximately like $0.95\angle(\x,\y)$ and $0.21+\frac{2}{3}\angle(\x,\y)$, respectively.}
\label{fig:angle_term_behav}
\end{figure}

Note that so far we assumed that $K$ is normalized and 
lies on the sphere $\mathbb{S}^{n-1}$.
Given that the data at the input of the network lie on the sphere and we use the ReLU $\rho$ as the activation function, the transformation $\rho(\M\cdot)$ keeps the output data (approximately) on a sphere (with half the diameter, see \eqref{eq:rho_M_euclidean_bound_signle} in the proof of Theorem~\ref{thm:DNN_euclidean_layeri} in the sequel). Therefore, in this case the normalization requirement holds up to a small distortion throughout the layers. This adds a motivation for having a normalization stage at the output of each layer, which was shown to provide some gain in several DNN \cite{Krizhevsky12ImageNet, Chatfield14Return}.

Normalization, which is also useful in shallow representations \cite{Perronnin10Improving}, can be interpreted as a transformation making the inner products between the data vectors coincide with the cosines of the corresponding angles.
While the bounds we provide in this section do not require normalization, they show that the operations of each layer rely on the angles between the data points. 
 
The following two results relate the Euclidean and angular distances in the input of a given layer to the ones in its output. We denote by $\mathbb{B}_r^n \subset \RR{n}$ the Euclidean ball of radius $r$. 

\begin{figure*}[htb]
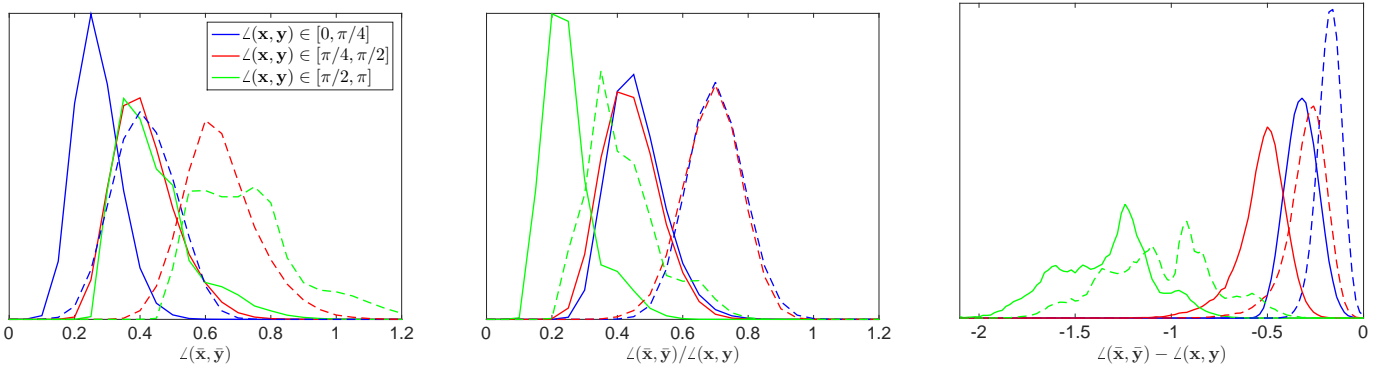

\begin{center}
{
\subfigure{\includegraphics[width=0.3\textwidth]{angle_rand_val16}\label{fig:ang_distort_imagnet16}}\hfill
\subfigure{\includegraphics[width=0.3\textwidth]{angle_rand_rat16}\label{fig:ang_distort_imagnet16_rat}}\hfill
\subfigure{\includegraphics[width=0.3\textwidth]{angle_rand_dif16}\label{fig:ang_distort_imagnet16_dif}}}
\end{center}
\caption{Left: Histogram of the angles in the output of the $8$-th (dashed-lines) and $16$-th (continuous-lines) layers of the ImageNet deep network for different input angle ranges.
Middle: Histogram of the ratio between the angles in the output of the $8$-th (dashed-line) and $16$-th (continuous-line) layers of this network    and the angles in its input for different input angle ranges.
Right: Histogram of the differences between the angles in the output of the $8$-th (dashed-line) and $16$-th (continuous-line) layers of this network    and the angles in its input for different input angles ranges.}
\label{fig:ang_distort_imagnet}
\end{figure*}

\begin{thm}
\label{thm:DNN_euclidean_layeri}
Let $\matr{M}$ be the linear operator applied at a given layer, $\rho$ (the ReLU) be the activation function, and $K \subset \mathbb{B}_1^n$ be the manifold of the data in the input layer. If ${\sqrt{m}}\matr{M} \in \RR{m \times n}$ is a random matrix with i.i.d. normally distributed entries and $m \ge C \delta^{-4}w(K)^4$, then with high probability 
\begin{eqnarray}
\label{eq:DNN_euclidean_bound_layeri}
&& \hspace{-0.6in} \Bigg\vert \norm{\rho(\matr{M}\vect{x}) - \rho(\matr{M}\vect{y})}_2^2  -  \\  \nonumber && 
 \left(\frac{1}{2}\norm{\x -\y }_2^2 + \norm{\x}_2\norm{\y}_2\dist(\x,\y)\right)
\Bigg\vert  \le \delta,
\end{eqnarray}
where $0 \le \angle (\x, \y) \triangleq \cos^{-1}\left(\frac{\x^T \y}{\norm{\x}_2\norm{\y}_2} \right) \le \pi$ and 
\begin{eqnarray}
\label{eq:euclead_add_term}
\dist(\x,\y) = \frac{1}{\pi}\Big( \sin\angle (\x, \y)  - 
 \angle (\x, \y)\cos\angle (\x, \y) 
 \Big).
\end{eqnarray} 
\end{thm}

\begin{thm}
\label{thm:DNN_angle_layeri}
Under the same conditions of Theorem~\ref{thm:DNN_euclidean_layeri} and $K \subset \mathbb{B}_1^n \setminus  \mathbb{B}_\beta^n$, where $\delta \ll \beta^2 < 1$,  with high probability
\begin{eqnarray}
\label{eq:DNN_angle_bound_layeri}
&& \hspace{-0.8in} \Bigg\vert \cos\angle\left( \rho(\matr{M}\vect{x}),  \rho(\matr{M}\vect{y})\right)  - 
 \\  \nonumber && 
 \left( \cos\angle (\x, \y)  + \dist(\x,\y) \right)
\Bigg\vert  \le \frac{15\delta}{\beta^2-2\delta}.
\end{eqnarray}
\end{thm}

 {\bf Remark 1} \emph{
As we have seen in Section~\ref{sec:model}, the assumption $m \ge C \delta^{-4}w(K)^2$ implies $m = O(k^2)$ if $K$ is a GMM and $m = O(k^2 \log L )$ if $K$ is generated by $k$-sparse representations in a dictionary $\matr{D} \in \RR{n \times L}$. As we shall see later in Theorem~\ref{thm:Covering_Stable_DNN} in Section~\ref{sec:ent_net}, it is enough to have the model assumption only at the data in the input of the DNN.
Finally, the quadratic relationship between $m$ and $w(K)^2$ might be improved.}

We leave the proof of these theorems to Appendices~\ref{sec:proof_euclid} and \ref{sec:proof_angle}, and dwell on their implications.

Note that if  $\dist(\x,\y)$
were equal to zero, then theorems~\ref{thm:DNN_euclidean_layeri} and \ref{thm:DNN_angle_layeri} would have stated that the distances and angles are preserved (in the former case, up to a factor of $0.5$\footnote{More specifically, we would have $\norm{\rho(\matr{M}\vect{x}) - \rho(\matr{M}\vect{y})}_2^2 \cong \frac{1}{2}\norm{\vect{x} - \vect{y}}_2^2$. Notice that this is the expectation of the distance $\norm{\rho(\matr{M}\vect{x} -\matr{M}\vect{y})}_2^2$.}) throughout the network.
As can be seen in Fig.~\ref{fig:cos_sin_term_behav}, $\dist(\x,\y)$ behaves approximately like $0.5(1-\cos\angle(\x,\y))$.
The larger is the angle $\angle (\x, \y)$, the larger is $\dist(\x,\y)$, and, consequently, also the Euclidean distance at the output. If the angle between $\x$ and $\y$ is close to zero (i.e., $\cos \angle (\x, \y)$ close to 1), $\dist(\x,\y)$ vanishes and therefore the Euclidean distance shrinks by half throughout the layers of the network. 
We emphasize that this is not in contradiction to Theorem \ref{thm:deep_net_stable} which guarantees approximately isometric embedding into the Hamming space. While the binarized output with the Hamming metric approximately preserves the input metric, the Euclidean metric on the raw output is distorted.

Considering this effect on the Euclidean distance, the smaller the angle between $\x$ and $\y$, the larger the distortion to this distance at the output of the layer, and the smaller the distance turns to be. On the other hand, the shrinkage of the distances is bounded as can be seen from the following corollary of Theorem~\ref{thm:DNN_euclidean_layeri}. 
\begin{cor}
\label{cor:DNN_euclidean_layeri_liph}
Under the same conditions of Theorem~\ref{thm:DNN_euclidean_layeri},  with high probability
\begin{eqnarray}
\label{eq:DNN_euclidean_liph_bound_layeri}
&& \hspace{-0.45in} \frac{1}{2}\norm{\x -\y }_2^2 - \delta \le \norm{\rho(\matr{M}\vect{x}) - \rho(\matr{M}\vect{y})}_2^2 
  \le \norm{\x -\y }_2^2 + \delta.
\end{eqnarray}
\end{cor}

The proof follows from the inequality of arithmetic and geometric means and the behavior of $\dist(\x,\y)$
(see Fig.~\ref{fig:cos_sin_term_behav}).
We conclude that DNN with random Gaussian weights preserve local structures in the manifold $K$ and enable decreasing distances between points away from it, a property much desired for classification.

The influence of the entire network on the angles is slightly different. Note that starting from the input of the second layer, all the vectors reside in the non-negative
 orthant. 
The cosine of the angles is translated from the range $[-1,1]$ in the first layer to the range $[0,1]$ in the subsequent second layers. In particular, the range $[-1,0]$ is translated to the range $[0,1/\pi]$, and in terms of the angle $\angle (\x, \y)$ from the range $[\pi/2, \pi]$ to $[\cos^{-1}(1/\pi), \pi/2]$.  
These angles shrink approximately by half, while the 
ones that are initially small remain approximately unchanged.

The action of the network preserves the order between the angles.  
Generally speaking, the network
affects the angles in the range $[0,\pi/2]$ in the same way. In particular, in the range  $[0, \pi/4]$ the angles in the output of the layer behave like $0.95\angle(\x,\y)$ and in the wider range $[0, \pi/2]$ they are bounded from below by  $0.8\angle(\x,\y)$ (see Fig.~\ref{fig:angle_term_behav}). 
Therefore, we conclude that the DNN distort the angles in the input manifold $K$ similarly and keep the general configuration of angles between the points.

To see that our theory captures the behavior of DNN endowed with pooling, we test how the angles change through the state-of-the-art $19$-layers deep network trained in \cite{Simonyan15VeryDeep} for the ImageNet dataset. We  randomly select  $3\cdot 10^4$ angles (pairs of data points) in the input of the network, partitioned to three equally-sized groups, each group corresponding to a one of the ranges  $[0,\pi/4]$, $[\pi/4, \pi/2]$ and $[\pi/2, \pi]$.
We test their behavior after applying eight and sixteen non-linear layers. 
The latter case corresponds to the part of the network excluding the fully connected layers.
We denote by $\bar\x$ the vector in the output of a layer corresponding to the input vector $\x$.
Fig.~\ref{fig:ang_distort_imagnet}  presents a histogram of the values of the angles $\angle(\bar\x, \bar\y)$ at the output of each of the layers for each of the three groups. It shows also the ratio $\angle(\bar\x, \bar\y)/\angle(\x,\y)$ and difference $\angle(\bar\x, \bar\y) - \angle(\x,\y)$, between the angles at the output of the layers and their original value at the input of the network.

As Theorem~\ref{thm:DNN_angle_layeri} predicts, the ratio $\angle(\bar{\x},\bar{\y}) / \angle(\x,\y)$ corresponding to $\angle(\x,\y) \in [\pi/2,\pi]$ is half the ratio corresponding to input angles in the range $[0, \pi/2]$. Furthermore, the ratios in the ranges $[0,\pi/4]$ and $[\pi/4, \pi/2]$ are approximately the same,
where in the range $[\pi/4,\pi/2]$ they are slightly larger. 
This is in line with Theorem~\ref{thm:DNN_angle_layeri} 
that claim that the angles in this range decrease approximately in the same rate, where for larger angles the shrink is slightly larger.
Also note that according to our theory the ratio corresponding to input angles in the range $[0,\pi/4]$ should behave on average like $0.95^q$, where $q$ is the number of layers. Indeed, for $q=8$, $0.95^8 = 0.66$ and for $q=16$, $0.95^{16} =0.44$; the centers of the histograms for the range $[0,\pi/4]$ are very close to these values. Notice that we have a similar behavior also for 
 the range $[\pi,\pi/2]$.
This is not surprising, as by looking at Fig.~\ref{fig:angle_term_behav}  one may observe that these angles also turn to be in the range that has the ratio $0.95$.
Remarkably, Fig.~\ref{fig:ang_distort_imagnet} demonstrates that 
the network keeps the order between the angles as Theorem~\ref{thm:DNN_angle_layeri} suggests. Notice that the shrinkage of the angles does not cause large angles to become smaller than other angles that were originally  significantly smaller than them.
 Moreover, small angles in the input remain small in the output as can be seen in Fig.~\ref{fig:ang_distort_imagnet}(right).

\begin{figure}[t]
\begin{center}
{
{\includegraphics[width=0.48\textwidth]{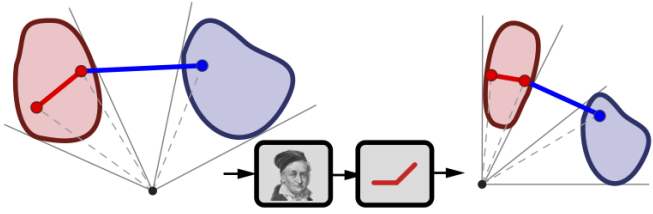}}}
\end{center}
\caption{Sketch of the distortion of two classes with distinguishable angle between them as obtained by one layer of DNN with random weights. These networks are suitable for separating this type of classes. Note that the distance between the blue and red points shrinks less than the distance between the red points as the angle between the latter is smaller.}
\label{fig:illustration}
\end{figure}

We sketch the distortion of two sets with distinguishable angle between them by one layer of DNN with random weights in Fig.~\ref{fig:illustration}. It can be observed that the distance between points with a smaller angle between them shrinks more than the distance between points with a larger angle between them.
Ideally, we would like this behavior, causing points belonging
 to the same class to stay closer to each other in the output of the network, compared to  points from different classes. 
However, random networks are not selective in this sense: if a point $\x$ forms the same angle with a point $\vect{z}$ from its class and with a  point $\vect{y}$ from another class, then their distance will be distorted approximately by an equal amount. Moreover, the separation caused by the network is dependent on the setting of the coordinate system origin with respect to which the angles are calculated. The location of the origin is dependent on the bias terms $\vect{b}$ (in this case each layer is of the form $\rho(\matr{M}\vect{x} + \vect{b})$), which are set to zero in the random networks here studied. These are learned by the training of the network, affecting the angles that cause the distortions of the Euclidean (and angular) distances. We demonstrate the effect of training in Section~\ref{sec:training}.

 \section{Embedding of the Entire Network}
\label{sec:ent_net}

In order to show that the results in sections~\ref{sec:model} and \ref{sec:dist_ang} also apply to the
\emph{entire} network and not only to one layer, we need to show that the Gaussian mean width does not grow significantly as the data propagate through the layers.
Instead of bounding the variation of the Gaussian mean width throughout the network, we bound the change in the covering number $N_{\epsilon}(K)$, i.e., the smallest number of $\ell_2$-balls of radius $\epsilon$ that cover $K$.
Having the bound on the covering number, we use Dudley's inequality \cite{Talagrand91Probability},
\begin{eqnarray}
\omega(K) \le C \int_0^\infty \sqrt{\log N_{\epsilon}(K)}d\epsilon,
\end{eqnarray}
to bound the Gaussian mean width. 

\begin{thm}
\label{thm:Covering_Stable_DNN}
Under the assumptions of Theorem~\ref{thm:deep_net_stable},
\begin{eqnarray}
N_{\epsilon}(f(\matr{M}K)) \le N_{\epsilon/\left(1+\frac{\omega(K)}{\sqrt{m}}\right)}\left(K \right). 
\end{eqnarray}
\end{thm}

{\it Proof:}  We divide the proof into two parts. In the first one, we consider the effect of the activation function $f$ on the size of the covering, while in the second we examine the effect of the linear transformation $\M$.
Starting with the activation function, let $\x_0 \in K$ be a center of a ball in the covering of $K$ and $\x \in K$ be a point that belongs to the ball of $\x_0$ of radius $\epsilon$, i.e., $\norm{\x - \x_0 }_2 \le \epsilon$.
It is not hard to see that since a semi-truncated linear
 activation function shrinks the data, then $\norm{f(\x) - f(\x_0)}_2 \le \norm{\x - \x_0}_2 \le \epsilon$ and therefore the size of the covering does not increase (but might decrease).

For the linear part we have that \cite[Theorem 1.4]{Klartag05Empirical} 
\begin{eqnarray}
\norm{\M\x - \M\x_0 }_2 \le \left(1+\frac{\omega(K)}{\sqrt{m}}\right) \norm{\x - \x_0}_2.
\end{eqnarray}
Therefore, after the linear operation each covering ball with initial radius $\epsilon$ is not bigger than $(1+\frac{\omega(K)}{\sqrt{m}})\epsilon$.
Since the activation function does not increase the size of the covering, we have that after a linear operation followed by an activation function, the size of the covering balls increases by a factor of $(1+\frac{\omega(K)}{\sqrt{m}})$. Therefore, the size of a covering with balls of radius $\epsilon$ of the output data $f(\M K)$ is bounded by the size of a covering with balls of radius $\epsilon/(1+\frac{\omega(K)}{\sqrt{m}})$.
 \hfill $\Box$ 

Theorem~\ref{thm:Covering_Stable_DNN}
generalizes the results in theorems~\ref{thm:deep_net_stable_rec}, \ref{thm:DNN_euclidean_layeri} and \ref{thm:DNN_angle_layeri} such that they can be used for the whole network: there exists an algorithm that recovers the input of the DNN from its output; the DNN as a whole distort the Euclidean distances based on the angels of the input of the network; and the angular distances smaller than $\pi$ are not altered significantly by the network. 

Note, however, that Theorem~\ref{thm:Covering_Stable_DNN} does not apply to Theorem~\ref{thm:deep_net_stable}.
In order to do that for the later, we need also a version of Theorem~\ref{thm:deep_net_stable} that guarantees a stable embedding using the same metric at the input and the output of a given layer, e.g., embedding from the Hamming cube to the Hamming cube or from $\mathbb{S}^{n-1}$ to $\mathbb{S}^{m-1}$. 
Indeed, we have exactly such a guarantee in
Corollary~\ref{cor:DNN_euclidean_layeri_liph} 
that implies stable embedding of the Euclidean distances in each layer of the network. Though this corollary focuses on the particular case of the ReLU, unlike Theorem~\ref{thm:deep_net_stable} that covers more general activation functions, it implies stability for the whole network in the Lipschitz sense, which is even stronger than stability in the Gromov-Hausdorff sense that we would get by having the generalization of Theorem~\ref{thm:deep_net_stable}.

As an implication of Theorem~\ref{thm:Covering_Stable_DNN}, consider low-dimensional data admitting  
 a Gaussian mixture model (GMM) with $L$ Gaussians of dimension $k$ or a $k$-sparse represention in a given dictionary with $L$ atoms. For GMM, the covering number is $N_\epsilon(K) = L\left(1 + \frac{2}{\epsilon} \right)^k$ for $\epsilon <1$ and $1$ otherwise (see \cite{Mendelson08Uniform}).
Therefore we have that $\omega(K) \le C\sqrt{k + \log{L}}$  and that at each layer the Gaussian mean width grows at most by a factor of $1+O\left(\frac{\sqrt{k} + \log{L}}{\sqrt{m}}\right)$. 
In the case of sparsely representable data, $N_{\epsilon}(K) = {L \choose k}\left(1 + \frac{2}{\epsilon} \right)^k$. By Stirling's approximation we have ${L \choose k} \le \left(\frac{e L}{k} \right)^k$ and therefore  $\omega(K) \le C\sqrt{k\log(L/k)}$.  Thus, at each layer the Gaussian mean width grows at most by a factor of $1+O\left(\frac{\sqrt{k\log(L/k)}}{\sqrt{m}}\right)$.

\section{Training Set Size}
\label{sec:meas_num}

An important question in deep learning is what is the amount of labeled  samples needed at training. 
Using Sudakov minoration \cite{Talagrand91Probability}, one can get an upper bound on the size of a covering of $K$, $N_\epsilon(K)$, which is the number of balls of radius $\epsilon$ that include all the points in $K$. 
We have demonstrated that networks with random Gaussian weights realize a stable embedding; consequently, if a network is trained using the screening technique by selecting the best among many networks generated with random weights as suggested in \cite{Pinto09High,Saxe11random, Beyond11Cox}, then the number of data points needed in order to guarantee that the network represents all the data is $O(\exp(\omega(K)^2 / \epsilon^2 ) )$. Since $\omega(K)^2$ is a proxy for the intrinsic data dimension as we have seen in the previous sections (see \cite{Plan13Robust} for more details), this bound formally predicts that the number of training points grows exponentially with the intrinsic dimension of the  data.

The exponential dependency is too pessimistic, as it is often possible to achieve a better bound on the required training sample size.
Indeed, the bound developed in \cite{Arora14Provable} requires much less samples. As the authors study the data recovery capability of an autoencoder, they assume that there exists a `ground-truth autoencoder'  generating the data. 
A combination of the data dimension bound here provided, with a prior on the relation of the data to a  deep network, should lead to a  better prediction of the number of needed training samples. In fact, we cannot refrain from drawing an analogy with the field of sparse representations of signals, where the combined use of the properties of the system with those of the input data led   to works that improved the bounds beyond the na\"{\i}ve manifold covering number 
(see \cite{Gribonval14Sparse} and references therein).

The following section presents such a combination,  by showing empirically that the purpose of training in DNN is to treat boundary points. This observation is likely to lead to a significant  reduction in the required size of the training data, and may also apply to active learning, where the training set is constructed adaptively.

\section{The Role of Training}
\label{sec:training}

The proof of Theorem~\ref{thm:DNN_euclidean_layeri} provides us with an insight on the role of training. One key property of the Gaussian distribution, which allows it to keep the ratio between the angles in the data, is its rotational invariance. 
The phase of a random Gaussian vector with i.i.d. entries is a random vector with a uniform distribution. Therefore, it does not prioritize one direction in the manifold over the other but treats all the same, leading to the behavior of the angles and distances throughout the net that we have described above.

In general, points within the same class would have small angles within them and points from distinct classes would have larger ones.
If this holds for all the points, then random Gaussian weights would be an ideal choice for the network parameters.
 However, as in practice this is rarely the case, an important role of the training would be to select in a smart way the separating hyper-planes induced by $\M$ in such a way that the angles between points from different classes are `penalized more' than the angles between the points in the same class.

Theorem~\ref{thm:DNN_euclidean_layeri} and its proof provide some  understanding of how this can be done by the learning process. Consider the expectation of the Euclidean distance between two points $\x$ and $\y$ at the output of a given layer. It reads as (the derivation appears in Appendix~\ref{sec:proof_euclid})
\begin{eqnarray}
&&\hspace{-0.25in} \mathbb{E}[\norm{\rho(\M\x) - \rho(\M\y)}_2^2]= \frac{1}{2}\norm{\x}_2^2 + \frac{1}{2}\norm{\y}_2^2  \\ \nonumber && ~~
+\frac{\norm{\x}_2\norm{\y}_2}{\pi} \int_0^{\pi - \angle(\x,\y) } \sin(\theta)\sin(\theta + \angle(\x,\y)) d\theta.
\end{eqnarray} 
Note that the integration in this formula is done uniformly over the interval $[0, \pi - \angle(\x,\y)]$, which contains the range of directions that have simultaneously positive inner products  with $\x$ and $\y$. With learning, we have the ability to pick the angle $\theta$ that maximizes/minimizes the inner product based on whether $\x$ and $\y$ belong to the same class or to distinct classes and in this way increase/decrease their Euclidean distances at the output of the layer.
 
Optimizing over all the angles between all the pairs of the points is a hard problem.  
This  explains why random initialization is a good choice for DNN. As it is hard to find the optimal configuration that separates the classes on the manifold, it is desirable to start with a universal one that treats most of the angles and distances well, and then to correct it in the locations that result in classification errors.

\begin{figure}[bt]
\begin{center}
{
\subfigure[Inter-class Euclidean distance ratio]{\includegraphics[width=0.22\textwidth]{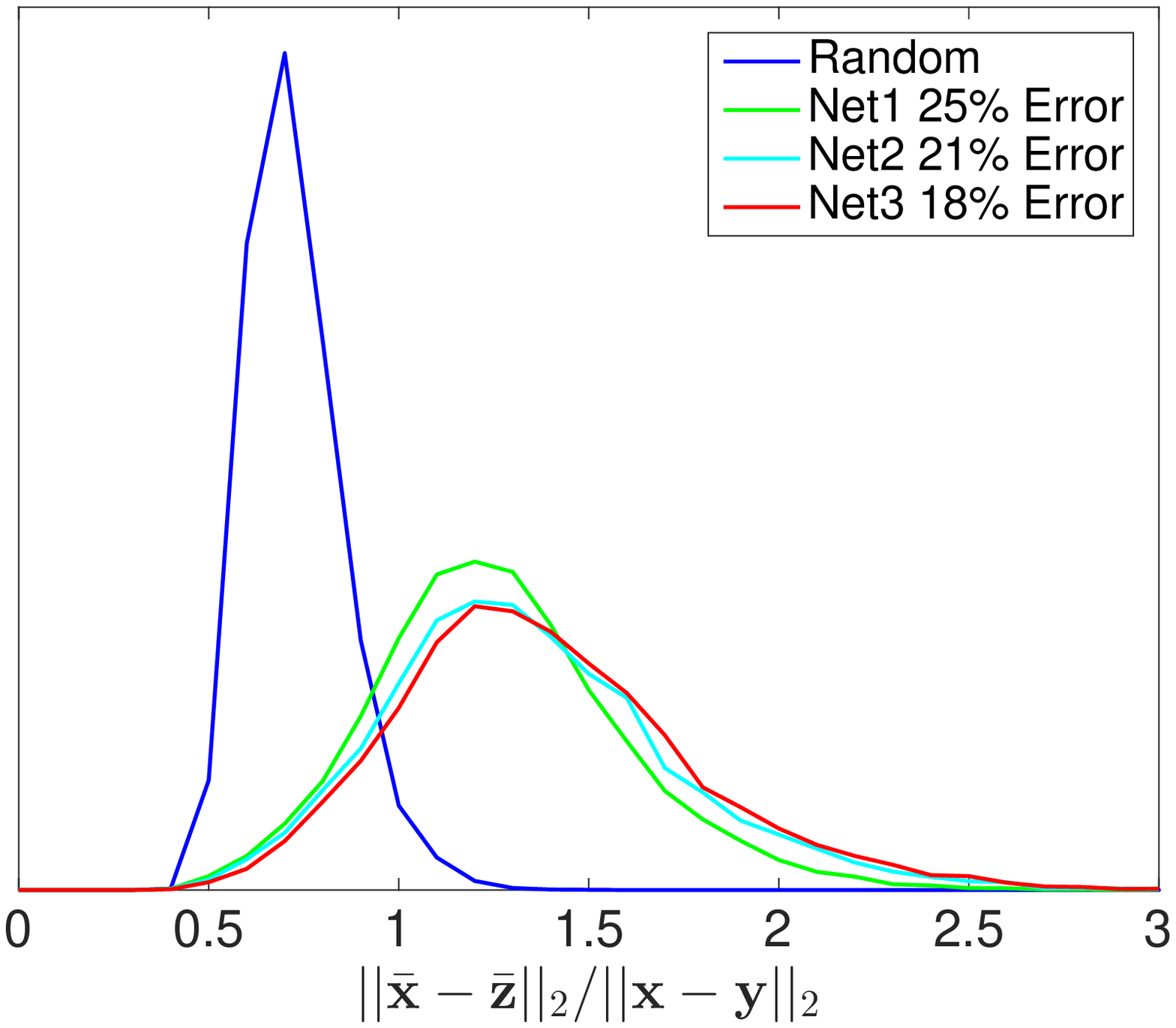}\label{fig:cifar_euclid_best_diff}}\hfill
\subfigure[Intra-class Euclidean distance ratio]{\includegraphics[width=0.22\textwidth]{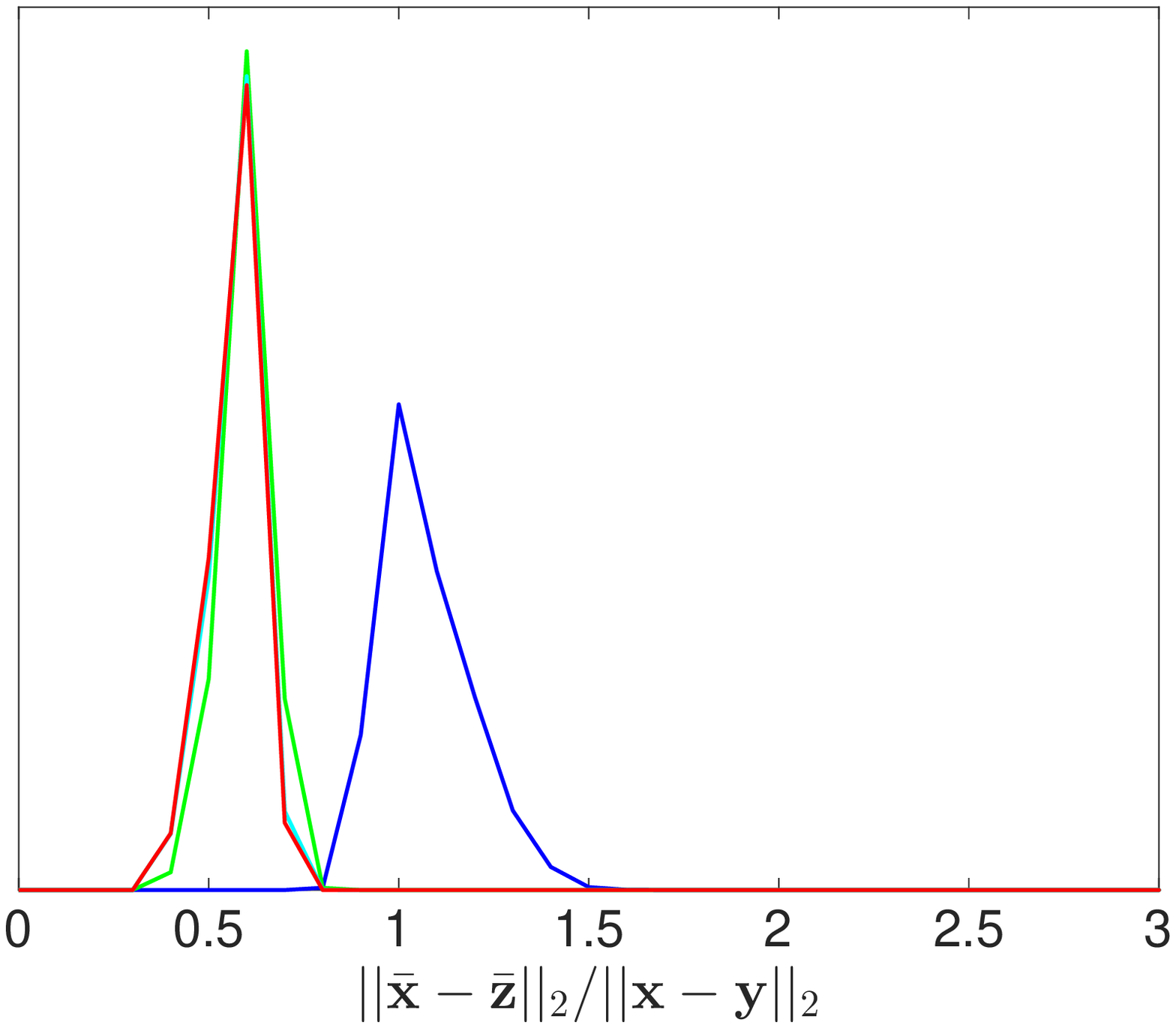}\label{fig:cifar_euclid_best_same}}
\subfigure[Inter-class angular distance ratio]{\includegraphics[width=0.22\textwidth]{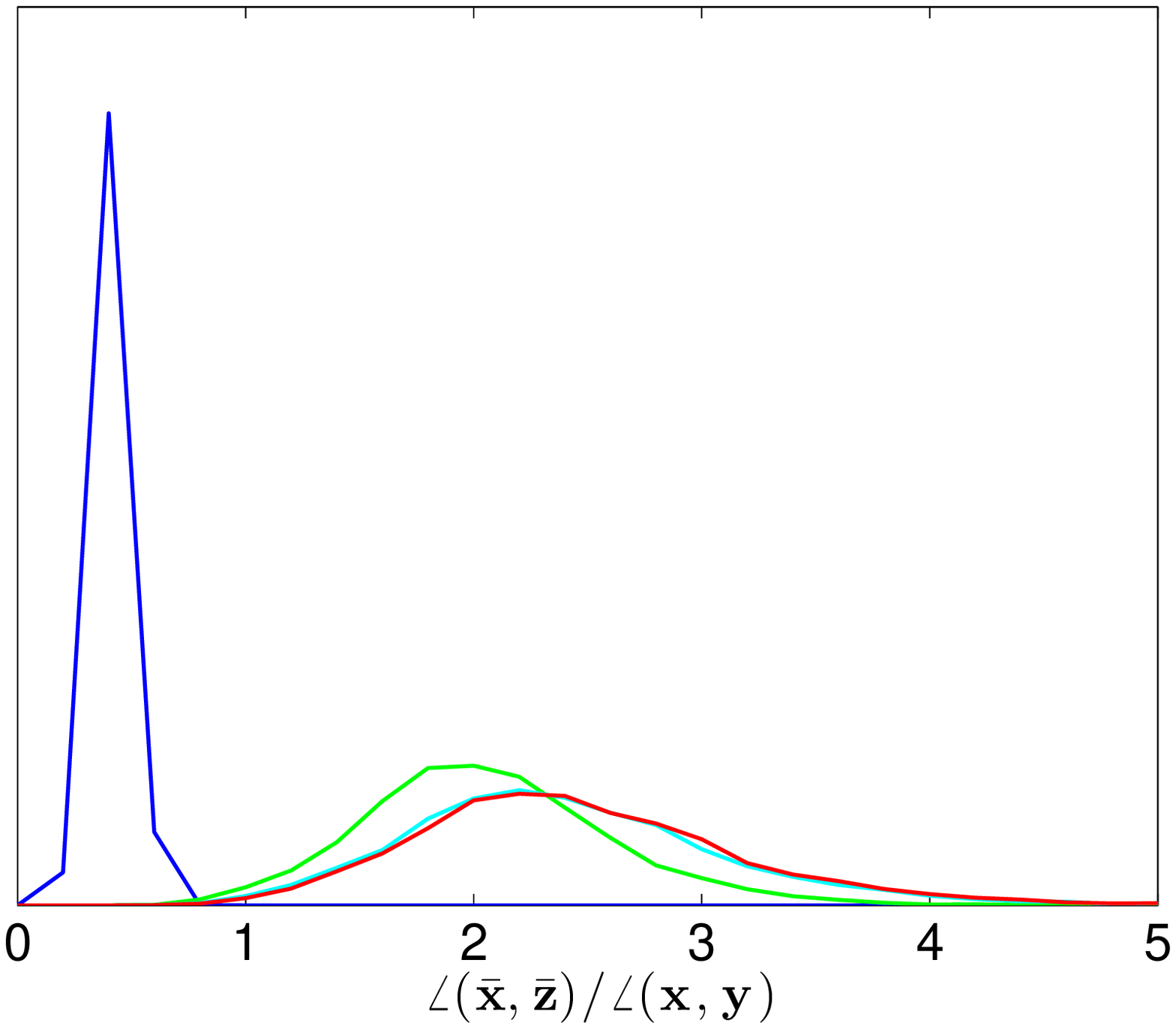}\label{fig:cifar_angle_best_diff}}\hfill
\subfigure[Intra-class angular distance ratio]{\includegraphics[width=0.22\textwidth]{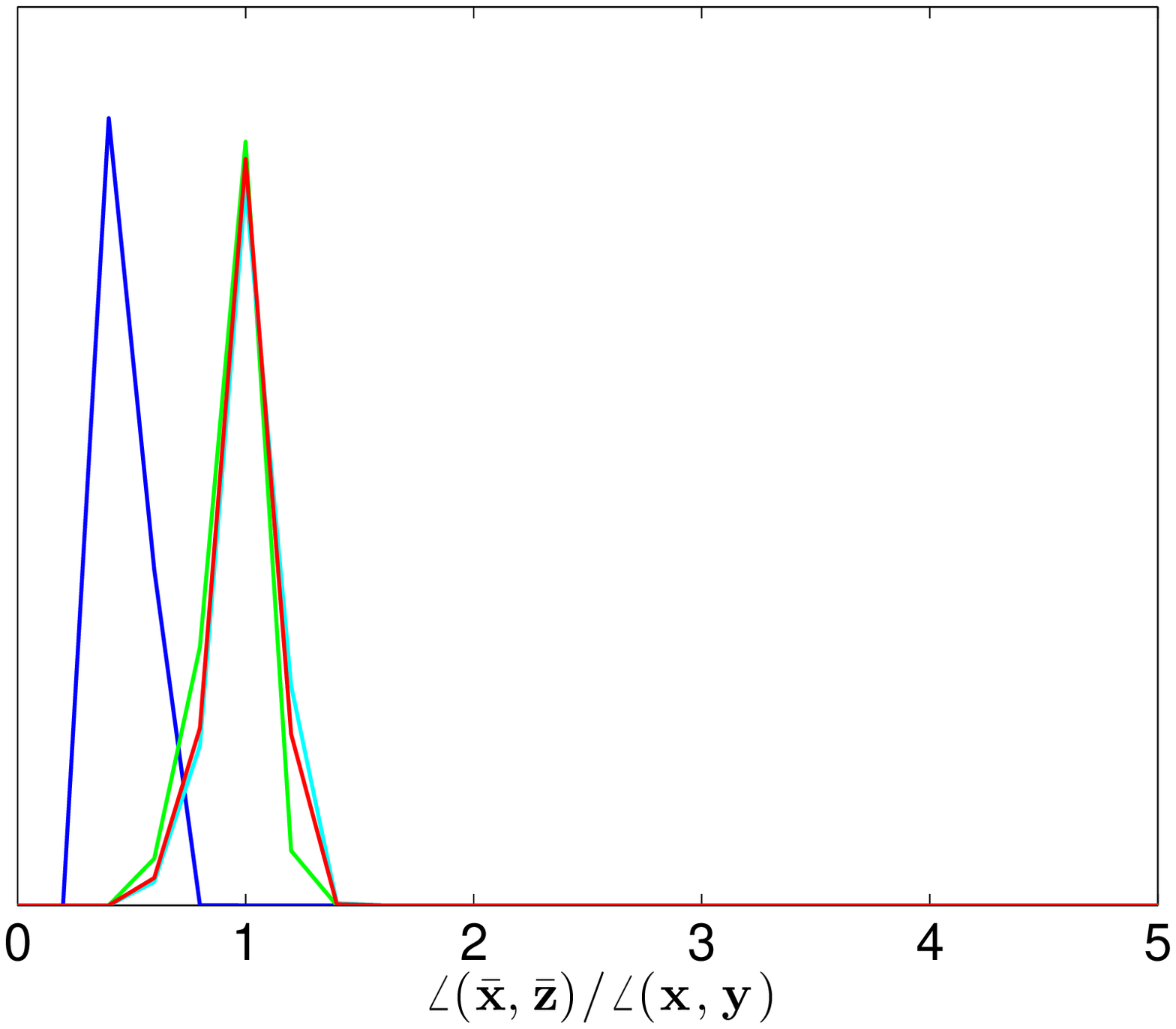}\label{fig:cifar_angle_best_same}}}
\end{center}
\caption{Ratios of \emph{closest} inter- (left) and \emph{farthest} intra-class (right) class Euclidean (top) and angular (bottom) distances for CIFAR-10. For each data point we calculate its Euclidean distance to the farthest point from its class and to the closest point not in its class, both at the input of the DNN and at the output of the last convolutional layer.
Then we compute the ratio between the two, i.e., if $\vect{x}$ is the point at input, $\vect{y}$ is its farthest point in class, $\bar{\vect{x}}$ is the point at the output, and $\bar{\vect{z}}$ is its farthest point from 
the same class (it should not necessarily be the output of $\vect{y}$), then we calculate $\frac{\norm{\bar{\vect{x}} - \bar{\vect{z}} }_2}{\norm{\vect{x} - \vect{y}}_2}$ and $\frac{\angle (\bar{\vect{x}},  \bar{\vect{z}}) }{\angle(\vect{x}, \vect{y})}$. We do the same for the distances between different classes, comparing the shortest Euclidean and angular distances.}
\label{fig:cifar_dist_best}
\end{figure}

\begin{figure}[bt]
\begin{center}
{
\subfigure[Inter-class Euclidean distance difference]{\includegraphics[width=0.23\textwidth]{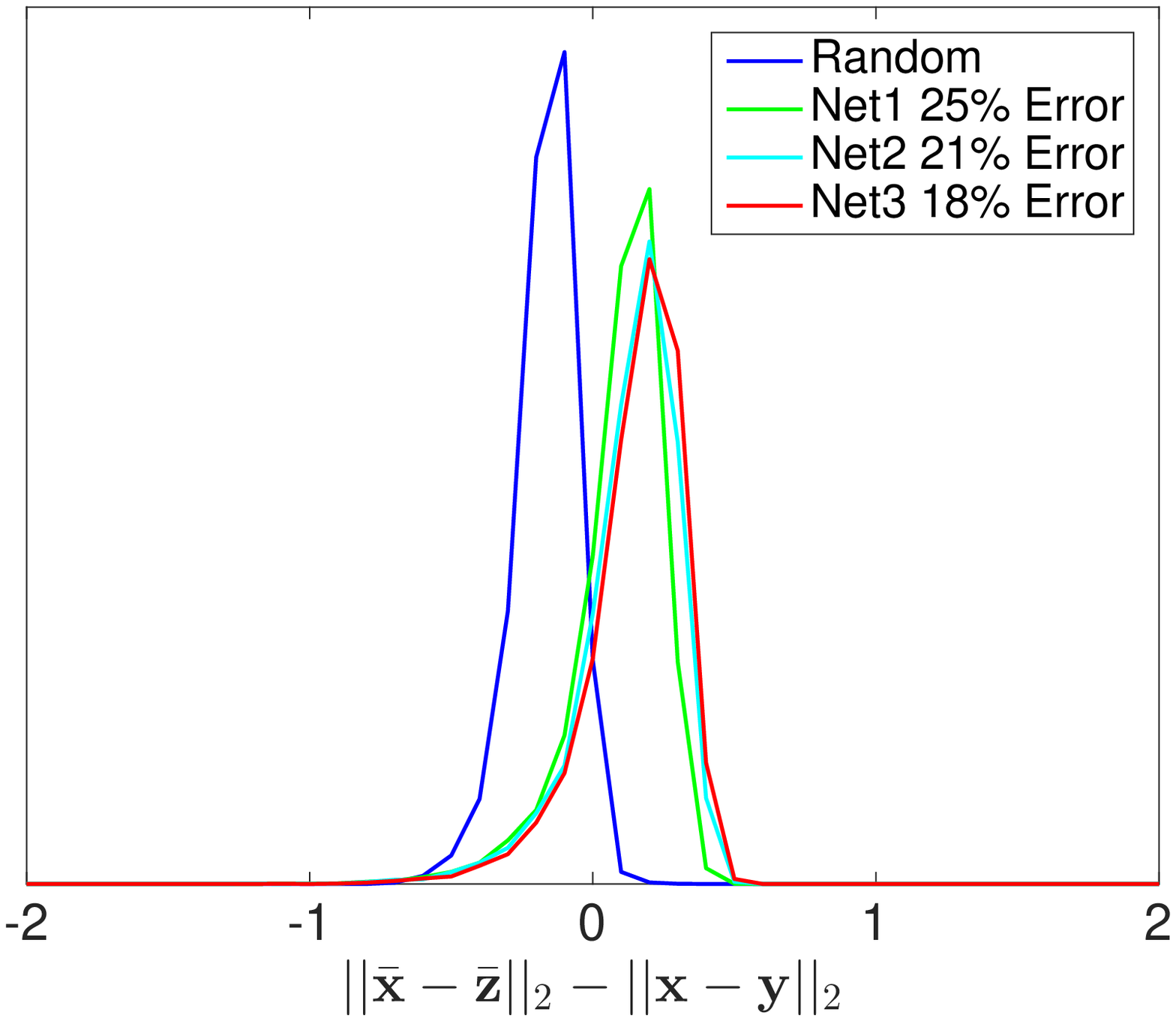}\label{fig:cifar_euclid_dif_best_diff}}\hfill
\subfigure[Intra-class Euclidean distance difference]{\includegraphics[width=0.23\textwidth]{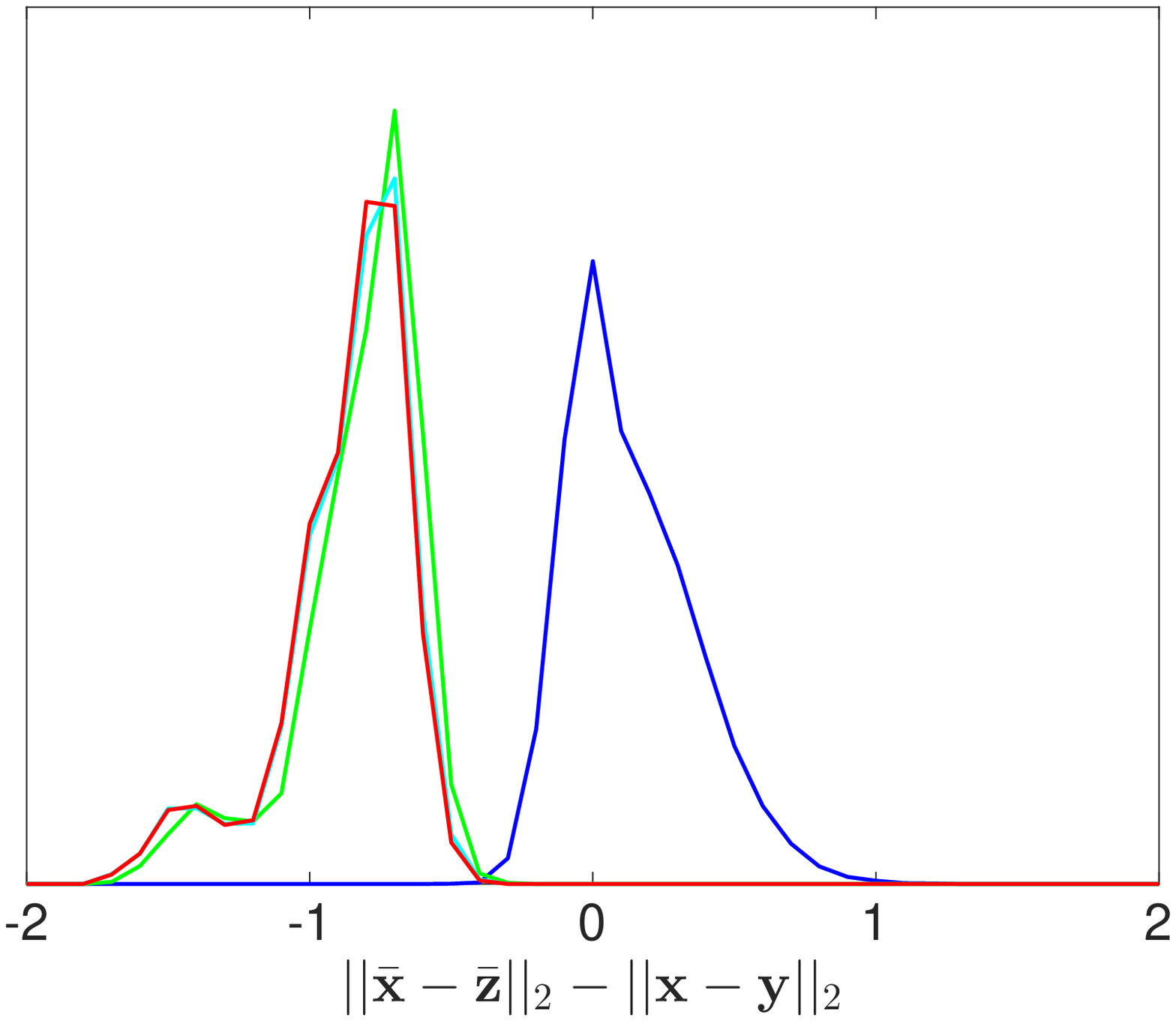}\label{fig:cifar_euclid_dif_best_same}}
\subfigure[Inter-class angular distance difference]{\includegraphics[width=0.23\textwidth]{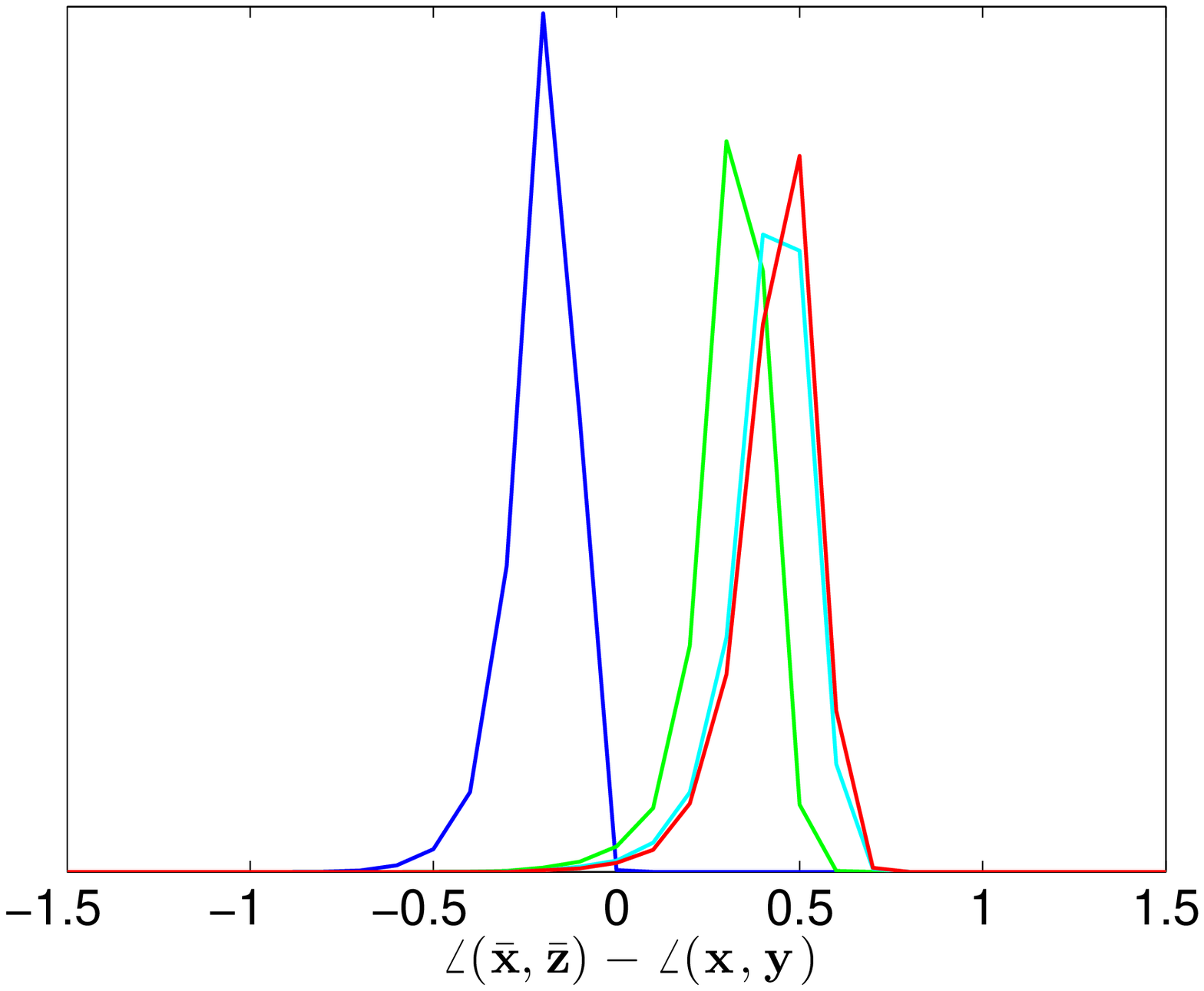}\label{fig:cifar_angle_dif_best_diff}}\hfill
\subfigure[Intra-class angular distance difference]{\includegraphics[width=0.23\textwidth]{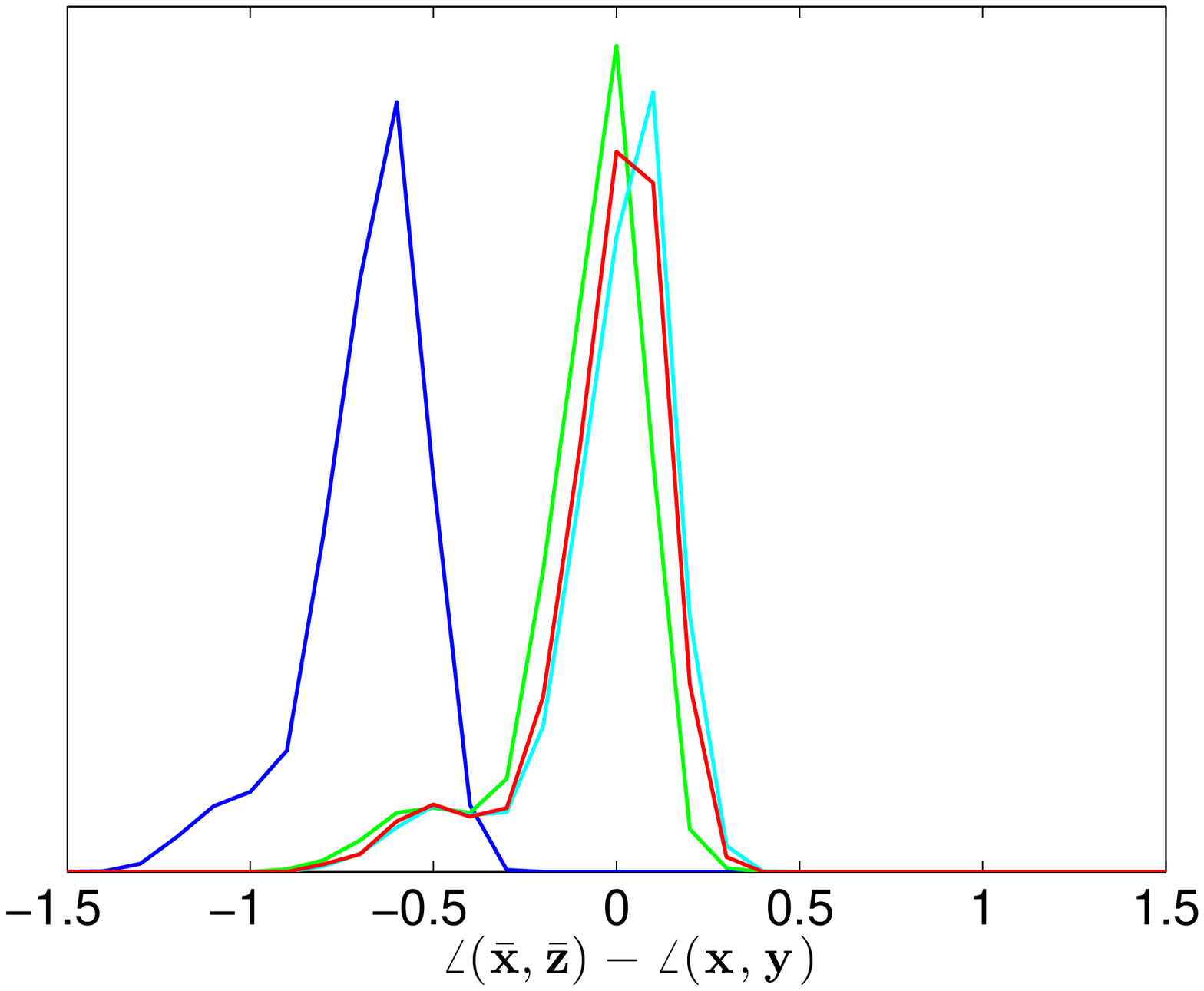}\label{fig:cifar_angle_dif_best_same}}}
\end{center}
\caption{Differences of \emph{closest} inter- (left) and \emph{farthest} intra- (right) class Euclidean (top) and angular (bottom) distances for CIFAR-10 (a counterpart of Fig.~\ref{fig:cifar_dist_best} with distance ratios replaced with differences). For each data point we calculate its Euclidean distance to the farthest point from its class and to the closest point not in its class, both at the input of the DNN and at the output of the last convolutional layer.
Then we compute the difference between the two, i.e., if $\vect{x}$ is the point at input, $\vect{y}$ is its farthest point in class, $\bar{\vect{x}}$ is the point at the output, and $\bar{\vect{z}}$ is its farthest point from 
the same class (it should not necessarily be the output of $\vect{y}$), then we calculate ${\norm{\bar{\vect{x}} - \bar{\vect{z}} }_2} - {\norm{\vect{x} - \vect{y}}_2}$ and ${\angle (\bar{\vect{x}},  \bar{\vect{z}}) } - {\angle(\vect{x}, \vect{y})}$. We do the same for the distances between different classes, comparing the shortest Euclidean and angular distances.}
\label{fig:cifar_dist_dif_best}
\end{figure}

\begin{figure}[bt]
\begin{center}
{
\subfigure[Inter-class Euclidean distance ratio]{\includegraphics[width=0.22\textwidth]{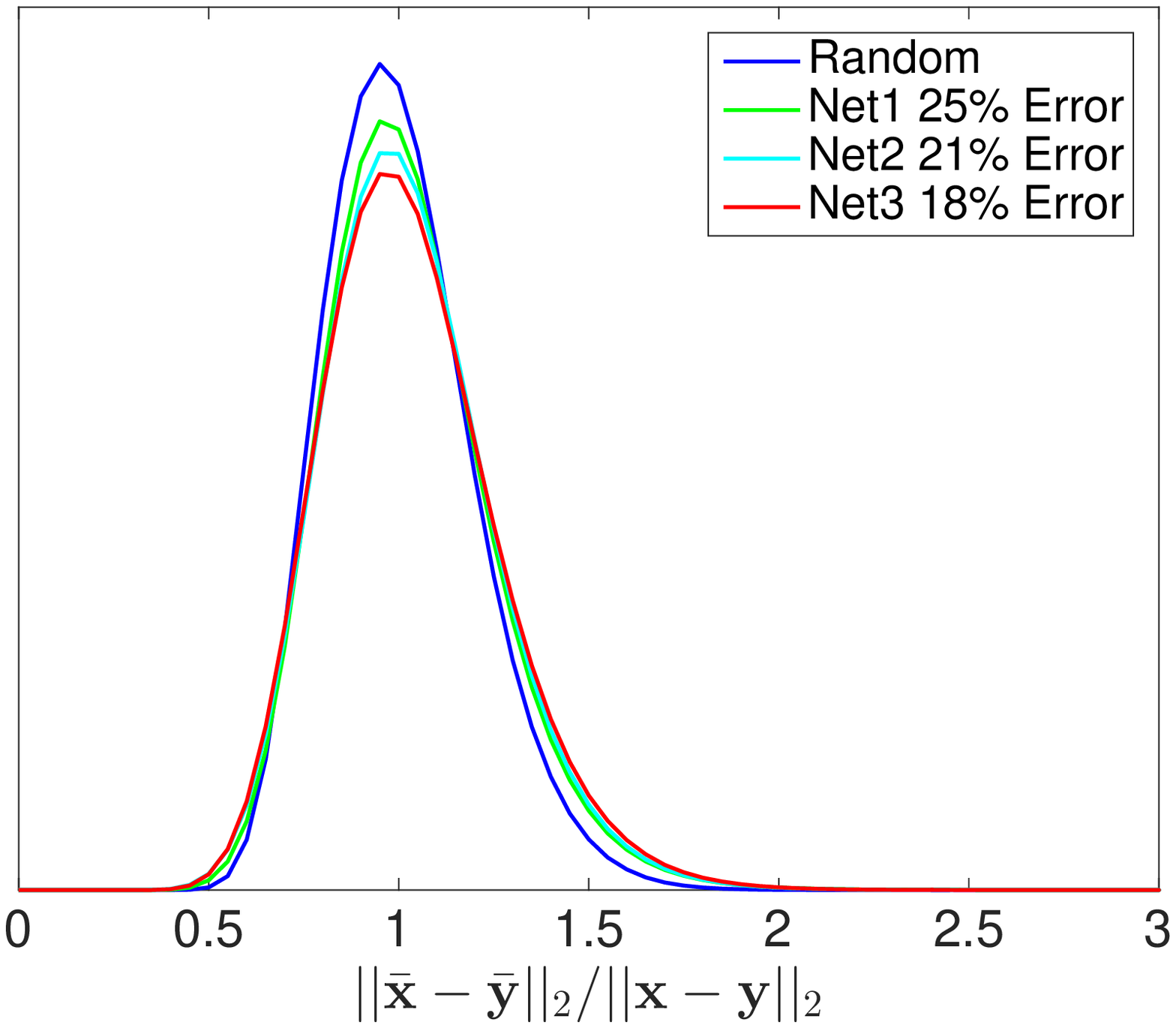}\label{fig:cifar_euclid_all_diff}}\hfill
\subfigure[Intra-class Euclidean distance ratio]{\includegraphics[width=0.22\textwidth]{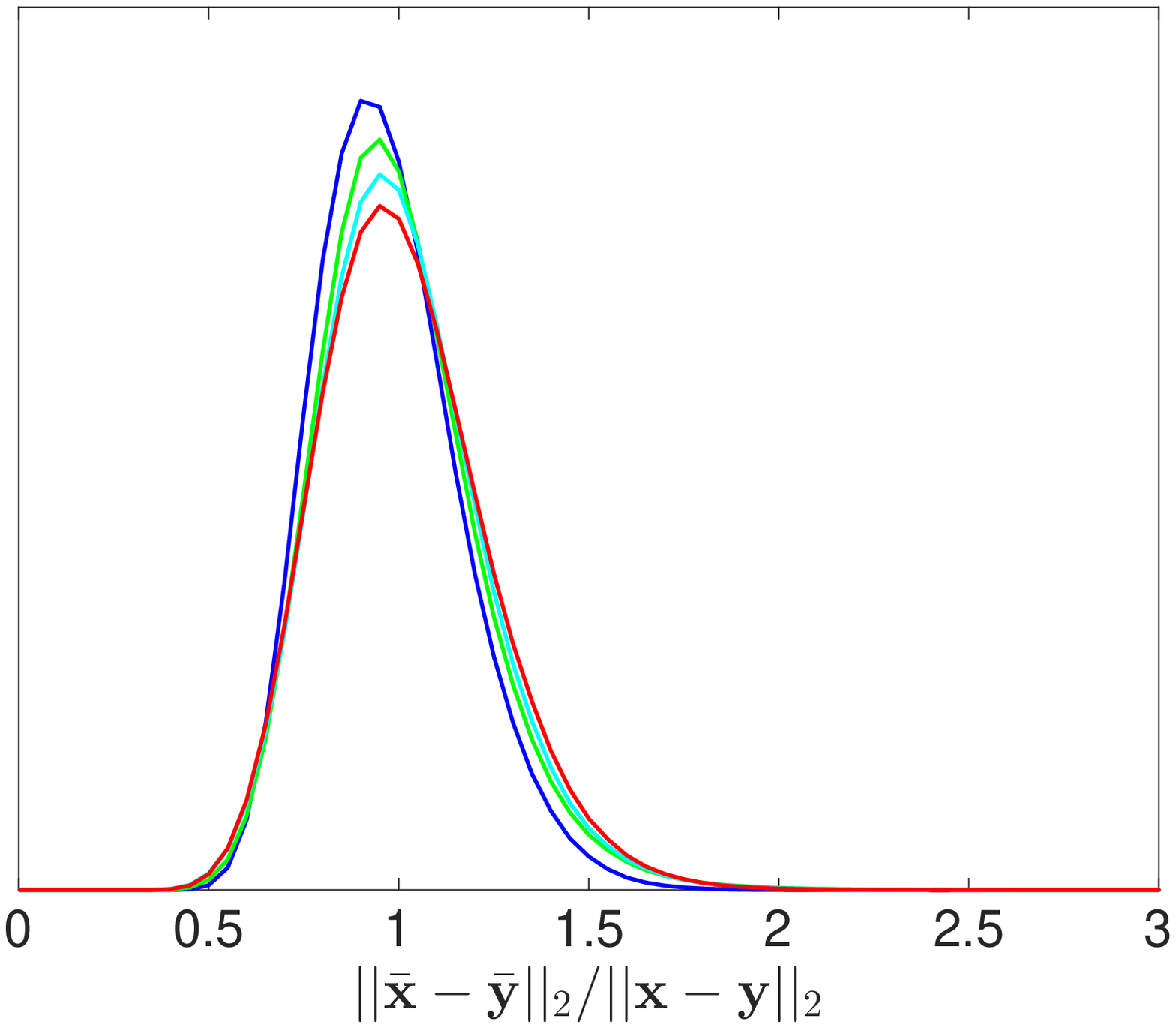}\label{fig:cifar_euclid_all_same}}
\subfigure[Inter-class angular distance ratio]{\includegraphics[width=0.22\textwidth]{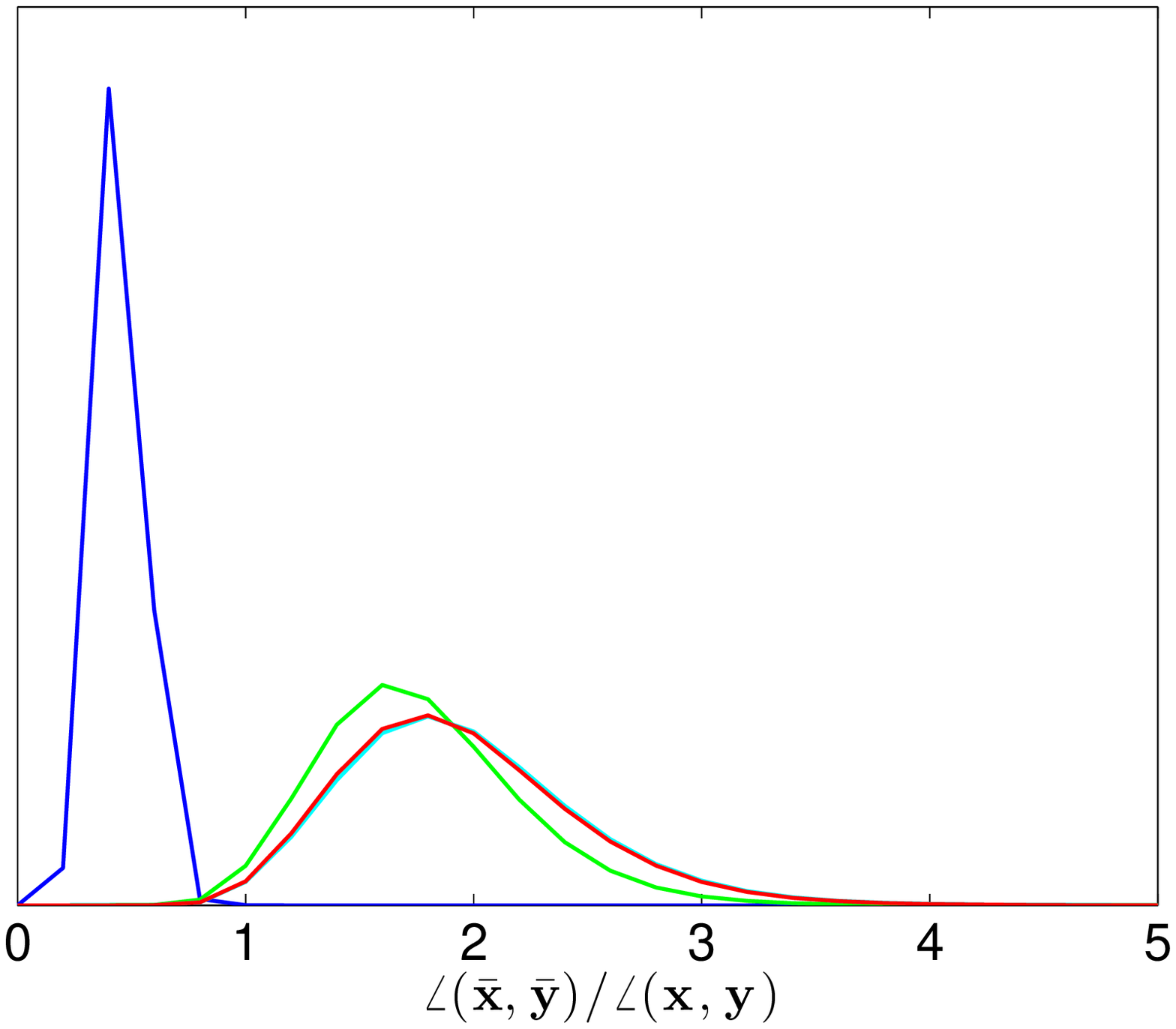}\label{fig:cifar_angle_all_diff}}\hfill
\subfigure[Intra-class angular distance ratio]{\includegraphics[width=0.22\textwidth]{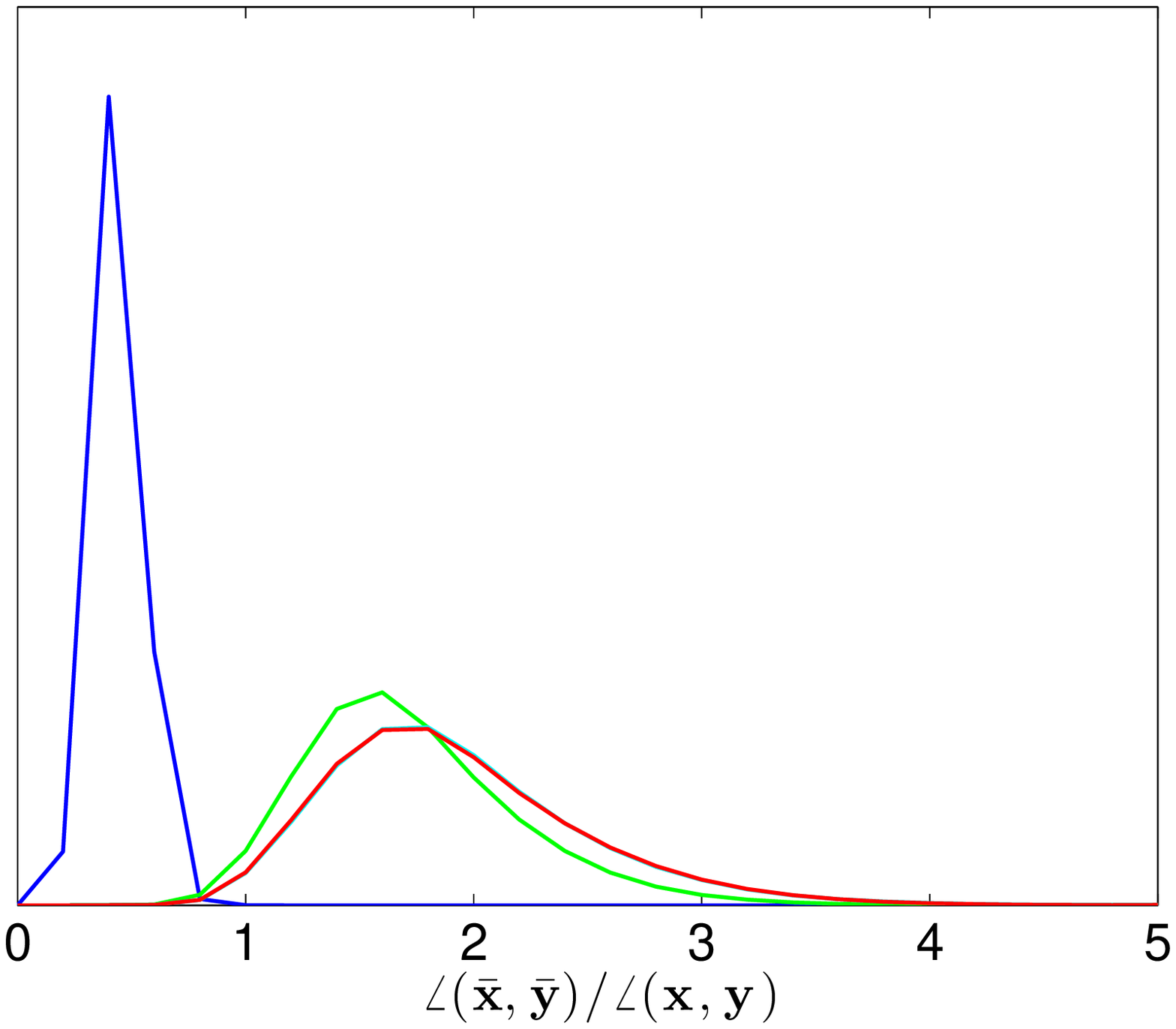}\label{fig:cifar_angle_all_same}}}
\end{center}
\caption{Ratios of inter- (left) and intra- (right) class Euclidean (top) and angular (bottom) distances between \emph{randomly selected points} for CIFAR-10.
We calculate the Euclidean distances between randomly selected pairs of data points from different classes (left) and from the same class (right), both at the input of the DNN and at the output of the last convolutional layer.
Then we compute the ratio between the two, i.e., for all pairs of points $(\vect{x}, \vect{y})$ in the input and their corresponding points $(\bar{\vect{x}}, \bar{\vect{y}})$ at the output we calculate $\frac{\norm{\bar{\vect{x}} - \bar{\vect{y}} }_2}{\norm{\vect{x} - \vect{y}}_2}$ and $\frac{\angle (\bar{\vect{x}},  \bar{\vect{y}}) }{\angle(\vect{x}, \vect{y})}$.}
\label{fig:cifar_dist_all}
\end{figure}

\begin{figure}[bt]
\begin{center}
{
\subfigure[Inter-class Euclidean distance difference]{\includegraphics[width=0.23\textwidth]{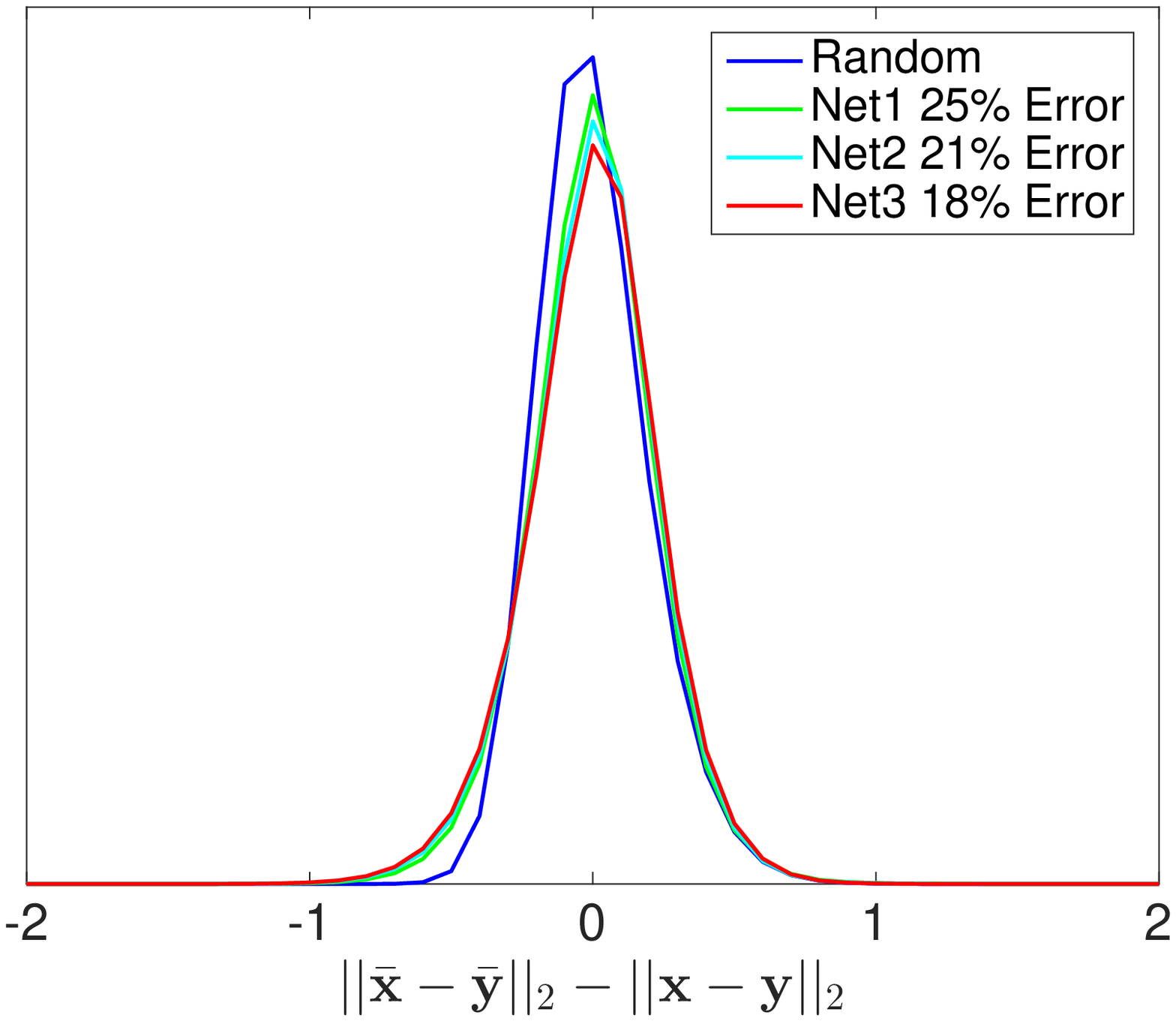}\label{fig:cifar_euclid_dif_all_diff}}\hfill
\subfigure[Intra-class Euclidean distance difference]{\includegraphics[width=0.23\textwidth]{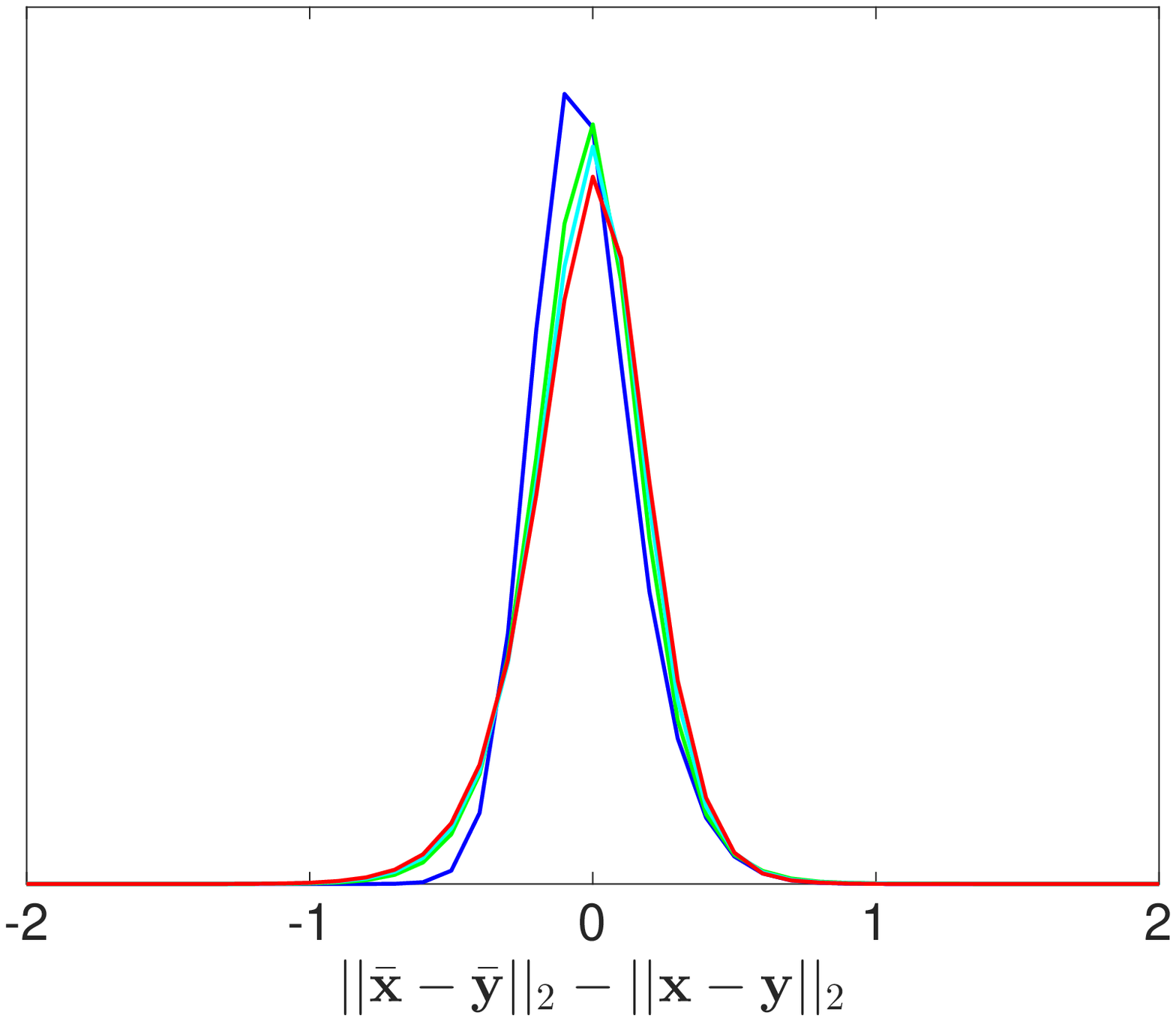}\label{fig:cifar_euclid_dif_all_same}}
\subfigure[Inter-class angular distance difference]{\includegraphics[width=0.23\textwidth]{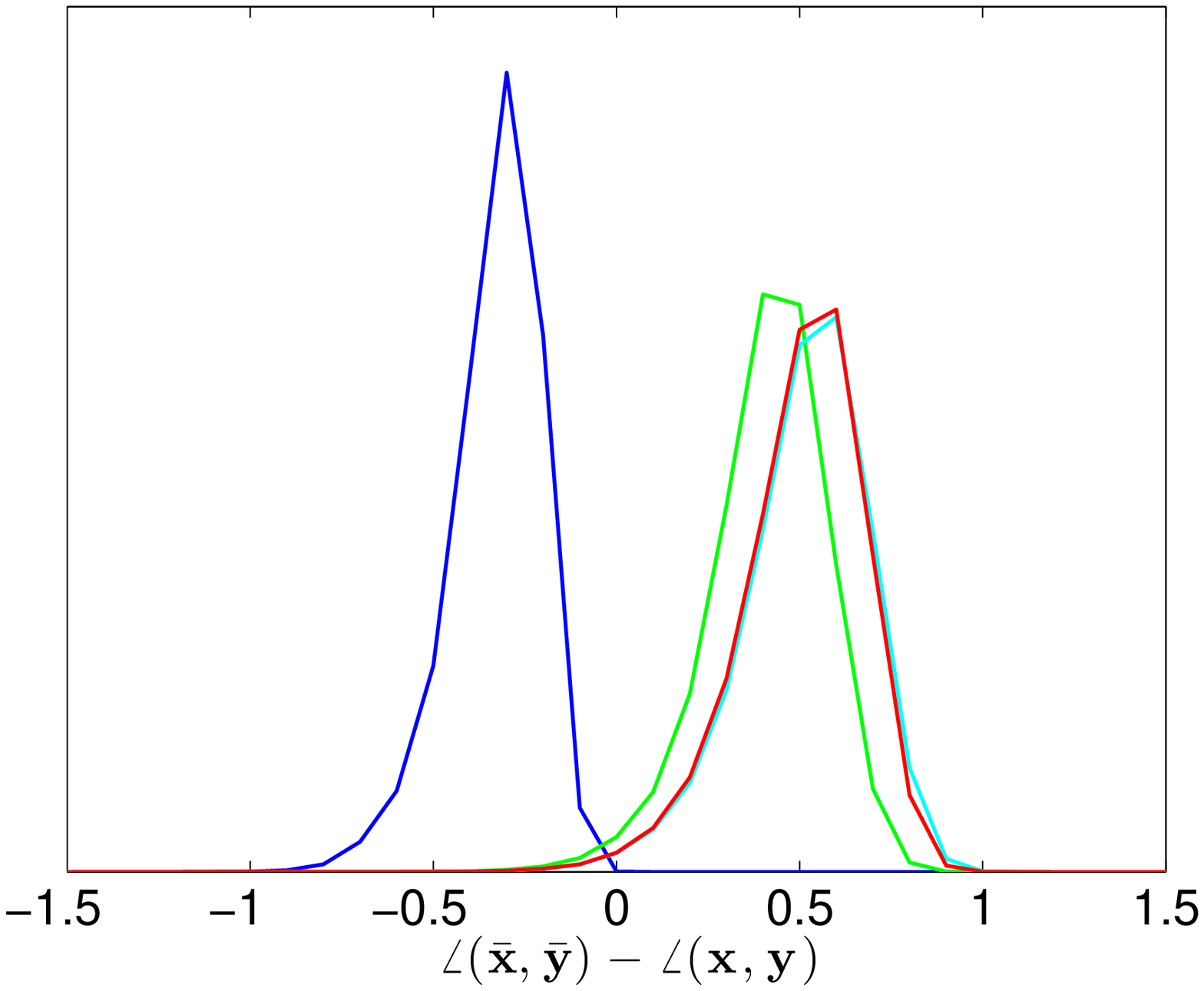}\label{fig:cifar_angle_dif_all_diff}}\hfill
\subfigure[Intra-class angular distance difference]{\includegraphics[width=0.23\textwidth]{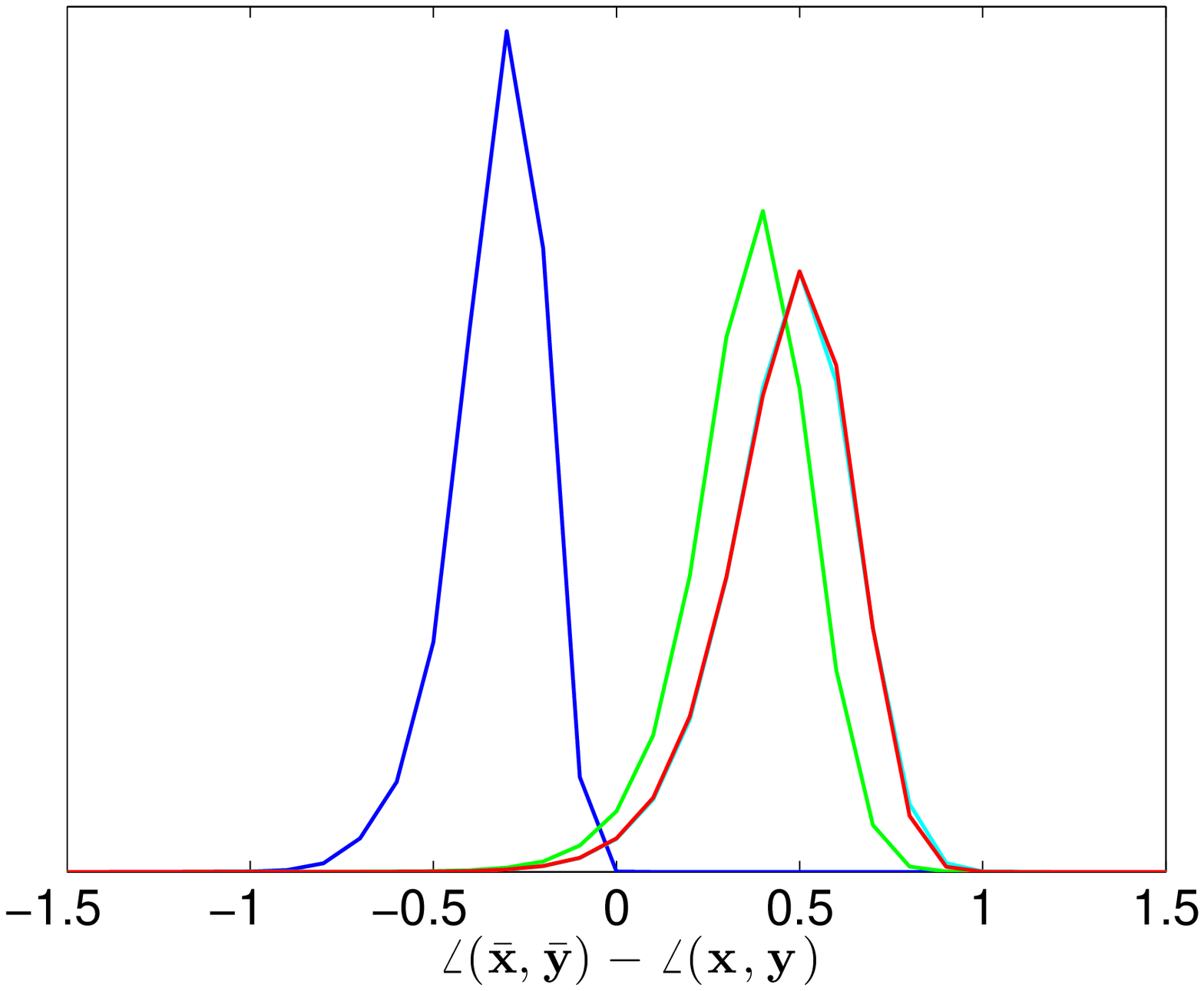}\label{fig:cifar_angle_dif_all_same}}}
\end{center}
\caption{Differences of inter- (left) and intra- (right) class Euclidean (top) and angular (bottom) distances between \emph{randomly selected points} for CIFAR-10 (a counterpart of Fig.~\ref{fig:cifar_dist_all} with distance ratios replaced with differences).
We calculate the Euclidean distances between randomly selected pairs of data points from different classes (left) and from the same class (right), both at the input of the DNN and at the output of the last convolutional layer.
Then we compute the difference between the two, i.e., for all pairs of points $(\vect{x}, \vect{y})$ in the input and their corresponding points $(\bar{\vect{x}}, \bar{\vect{y}})$ at the output we calculate ${\norm{\bar{\vect{x}} - \bar{\vect{y}} }_2} - {\norm{\vect{x} - \vect{y}}_2}$ and ${\angle (\bar{\vect{x}},  \bar{\vect{y}}) }-{\angle(\vect{x}, \vect{y})}$.}
\label{fig:cifar_dist_dif_all}
\end{figure}

%
%

\begin{figure}[bt]
\begin{center}
{
\subfigure[Inter-class Euclidean distance ratio]{\includegraphics[width=0.22\textwidth]{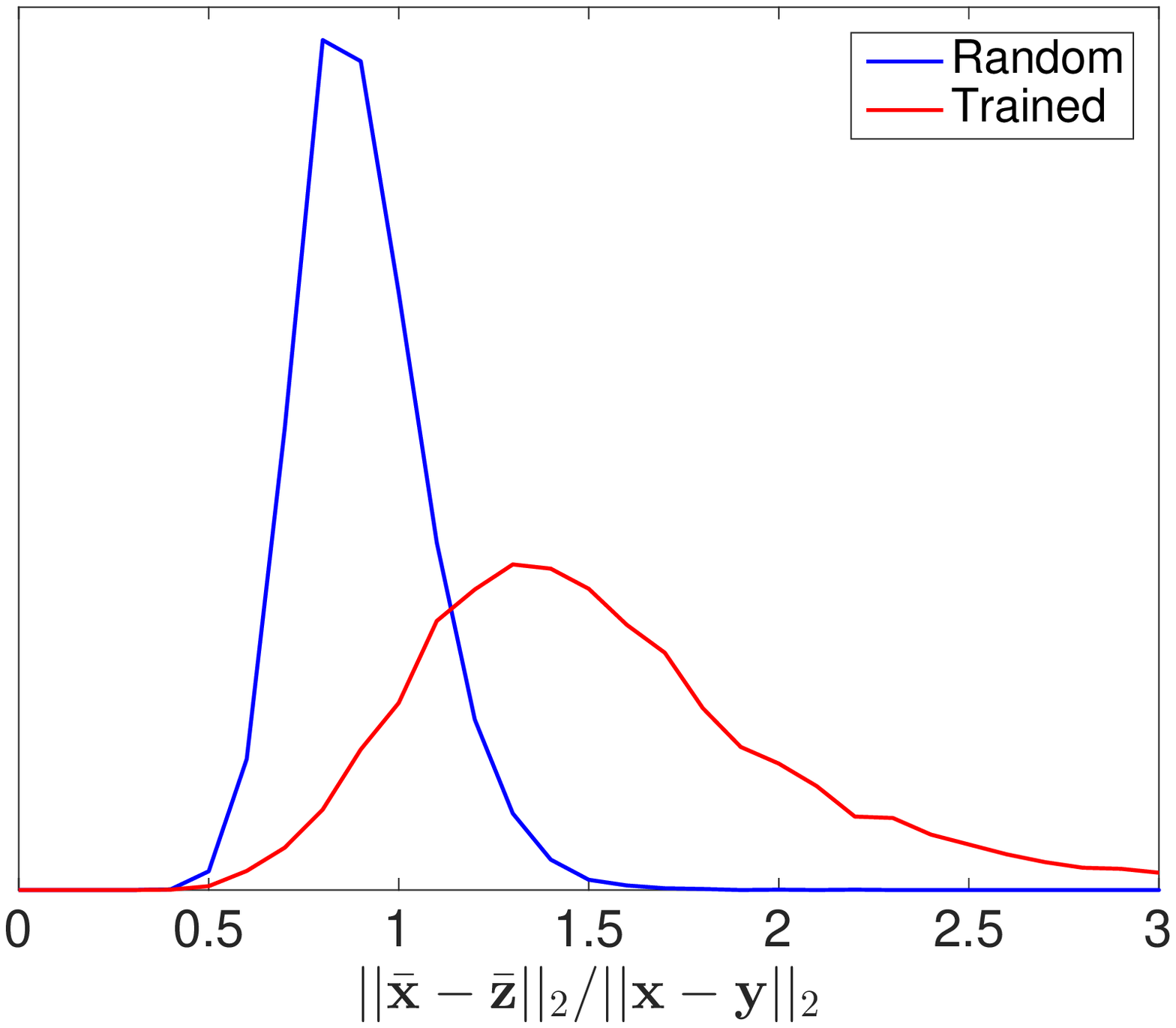}\label{fig:imagenet_euclid_best_diff}}\hfill
\subfigure[Intra-class Euclidean distance ratio]{\includegraphics[width=0.22\textwidth]{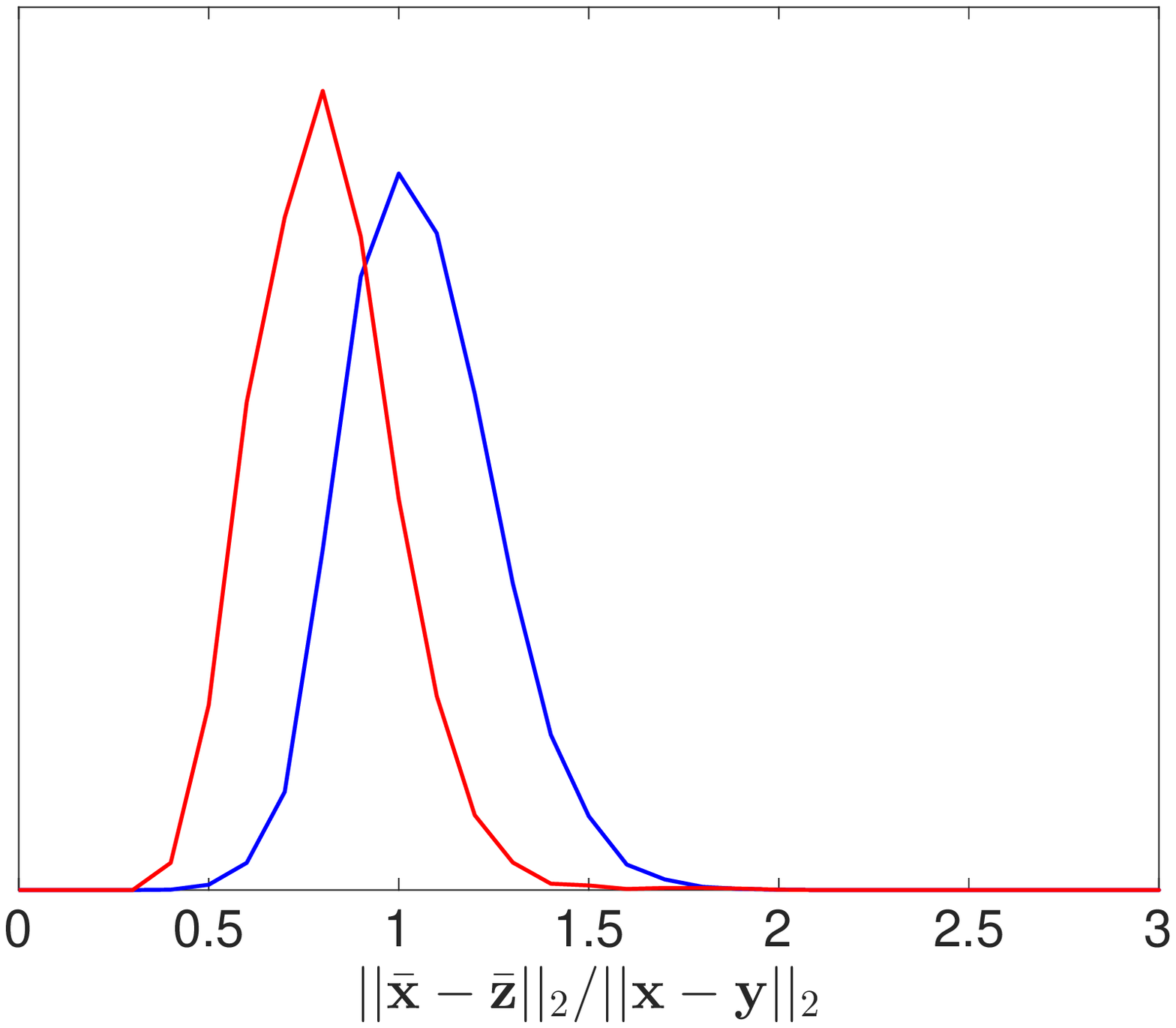}\label{fig:imagenet_euclid_best_same}}
\subfigure[Inter-class angular distance ratio]{\includegraphics[width=0.22\textwidth]{imagenet_angle_best_diff}\label{fig:imagenet_angle_best_diff}}\hfill
\subfigure[Intra-class angular distance ratio]{\includegraphics[width=0.22\textwidth]{imagenet_angle_best_same}\label{fig:imagenet_angle_best_same}}
}
\end{center}
\caption{Ratios of \emph{closest} inter- (left) and \emph{farthest} intra- (right) class Euclidean (top) and angular (bottom) distances for ImageNet. For each data point we calculate its Euclidean distance to the farthest point from its class and to the closest point not in its class, both at the input of the DNN and at the output of the last convolutional layer.
Then we compute the ratio between the two, i.e., if $\vect{x}$ is the point at input, $\vect{y}$ is its farthest point in class, $\bar{\vect{x}}$ is the point at the output, and $\bar{\vect{z}}$ is its farthest point from 
the same class (it should not necessarily be the output of $\vect{y}$), then we calculate $\frac{\norm{\bar{\vect{x}} - \bar{\vect{z}} }_2}{\norm{\vect{x} - \vect{y}}_2}$ and $\frac{\angle (\bar{\vect{x}},  \bar{\vect{z}}) }{\angle(\vect{x}, \vect{y})}$. We do the same for the distances between different classes, comparing the shortest Euclidean and angular distances.}
\label{fig:imagenet_dist_best}
\end{figure}

\begin{figure}[bt]
\begin{center}
{
\hspace{0.1in}\subfigure[Inter-class Euclidean distance difference]{\includegraphics[width=0.22\textwidth]{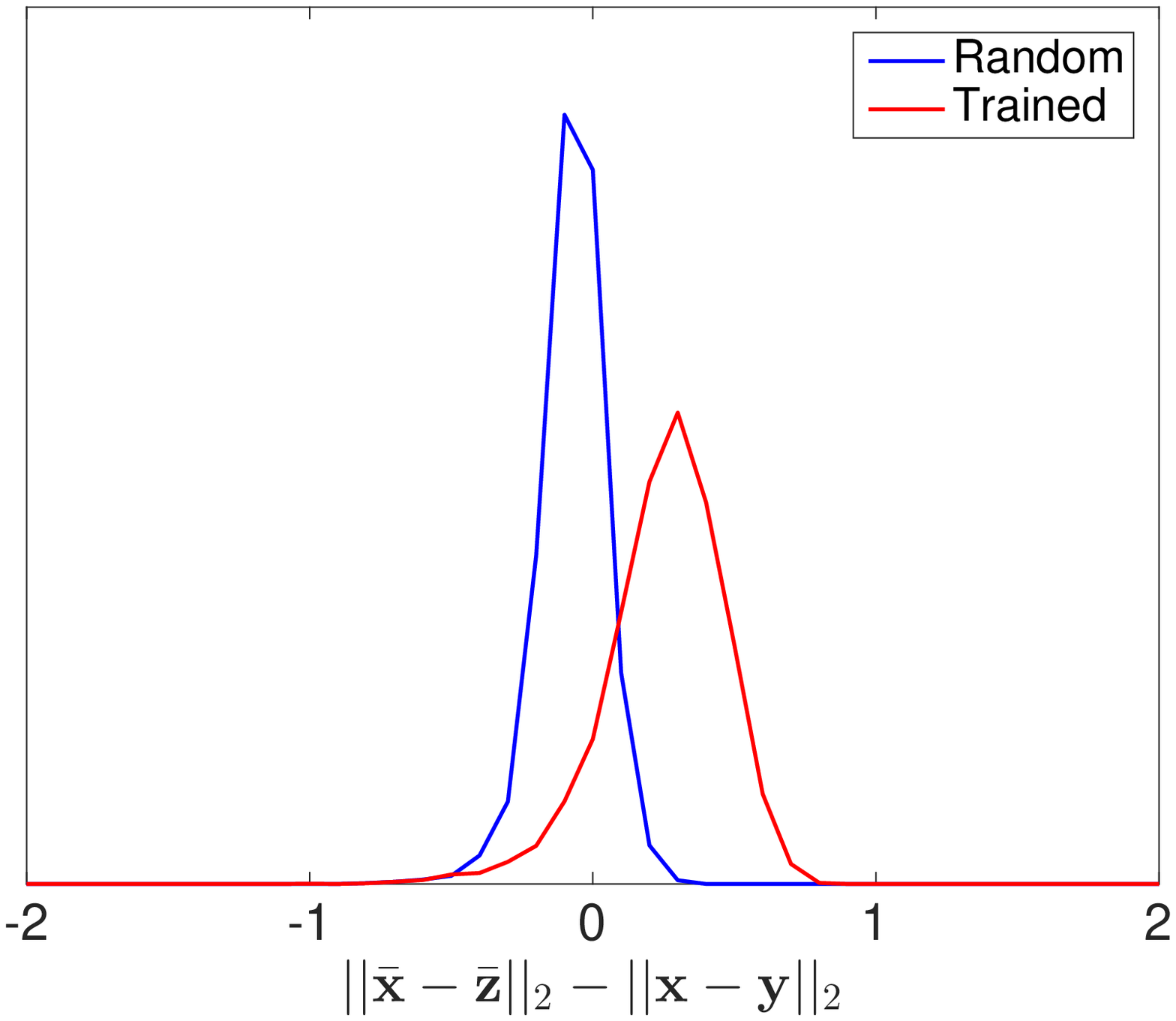}\label{fig:imagenet_euclid_dif_best_diff}}\hfill
\subfigure[Intra-class Euclidean distance difference]{\includegraphics[width=0.22\textwidth]{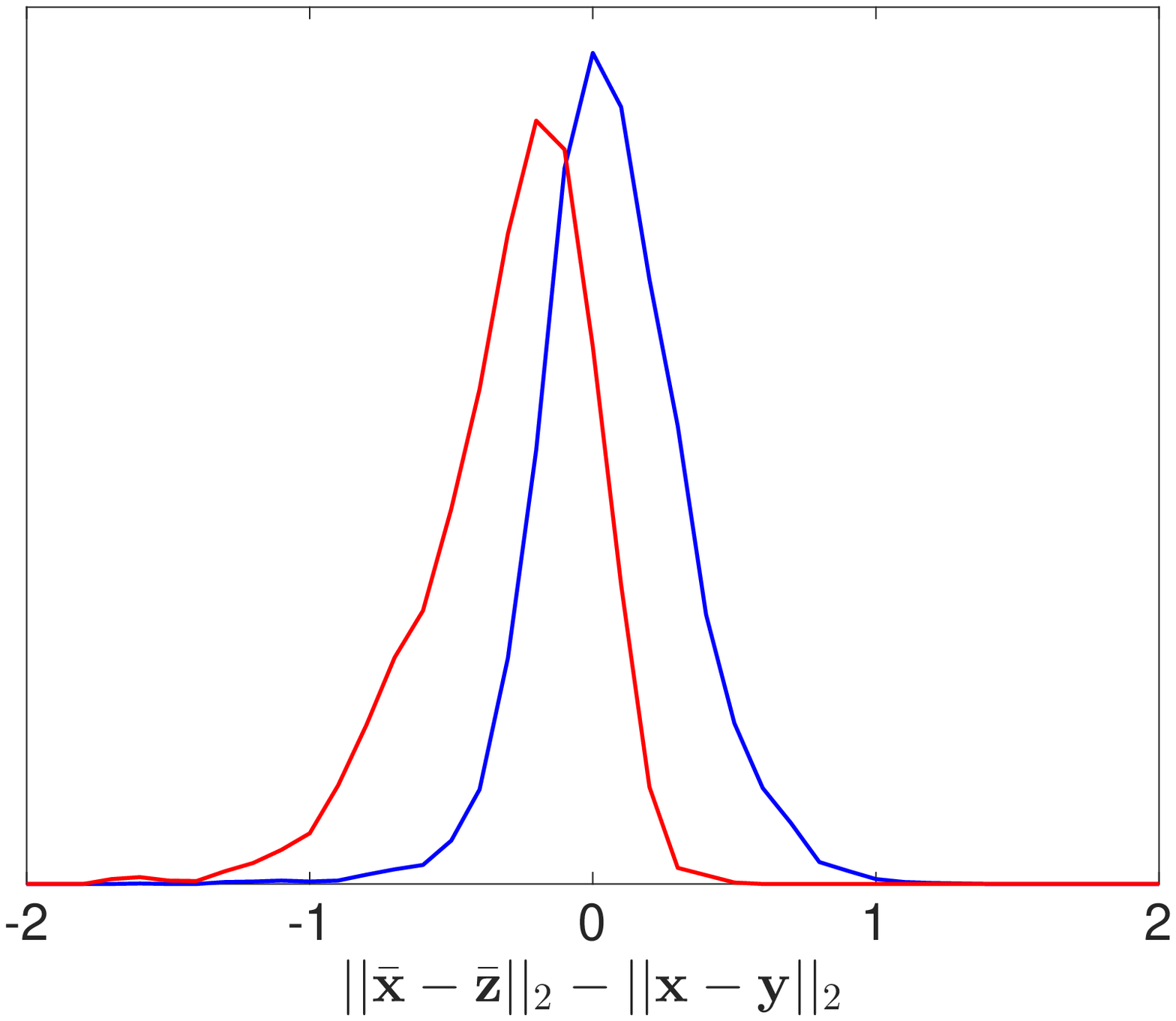}\label{fig:imagenet_euclid_dif_best_same}}
\hspace{0.08in}\subfigure[Inter-class angular distance difference]{\includegraphics[width=0.23\textwidth]{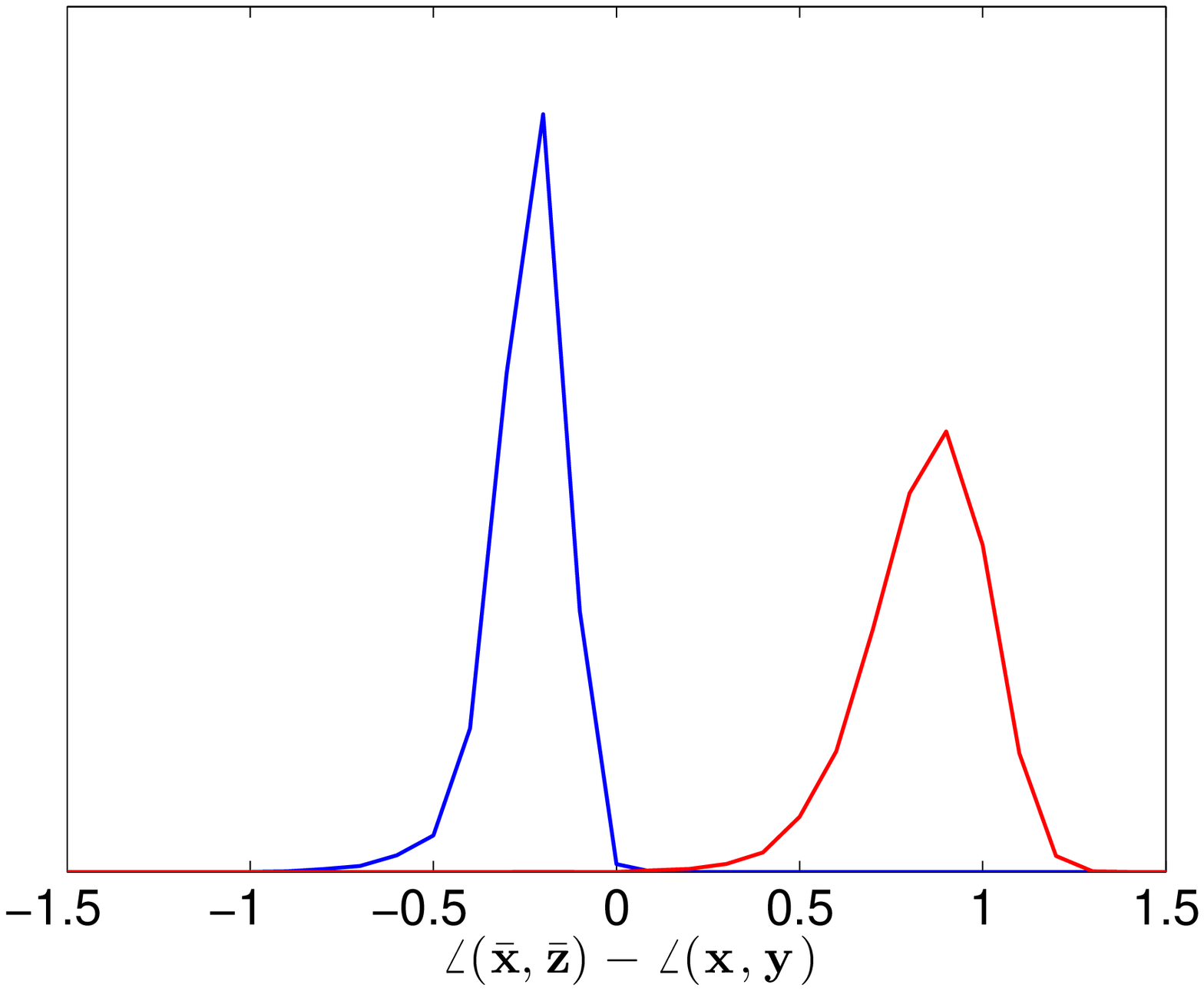}\label{fig:imagenet_angle_dif_best_diff}}\hfill
\subfigure[Intra-class angular distance difference]{\includegraphics[width=0.23\textwidth]{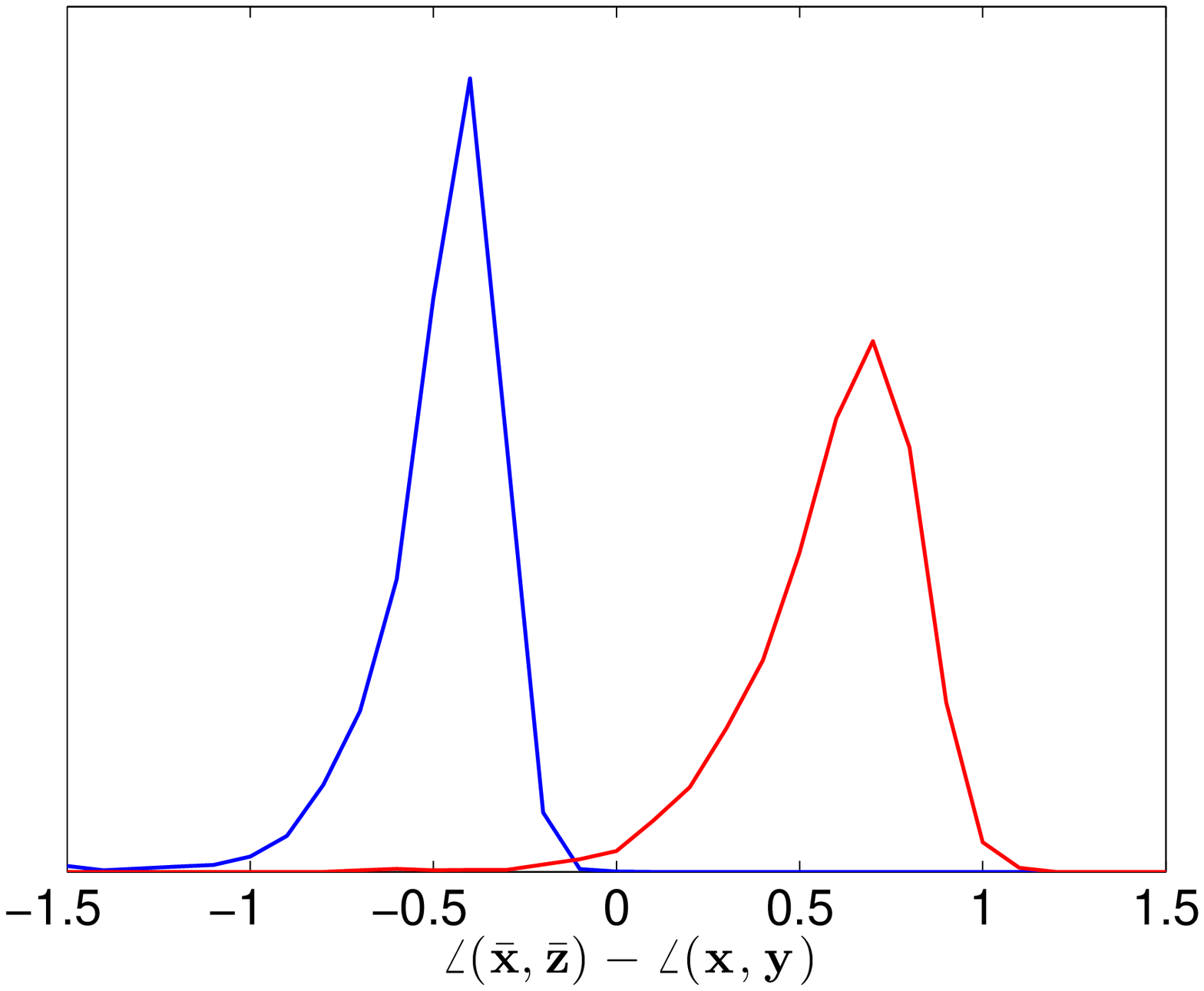}\label{fig:imagenet_angle_dif_best_same}}
}
\end{center}
\caption{Differences of \emph{closest} inter- (left) and \emph{farthest} intra- (right) class Euclidean (top) and angular (bottom) distances for ImageNet (a counterpart of Fig.~\ref{fig:imagenet_dist_best} with distance ratios replaced with differences). For each data point we calculate its Euclidean distance to the farthest point from its class and to the closest point not in its class, both at the input of the DNN and at the output of the last convolutional layer.
Then we compute the difference between the two, i.e., if $\vect{x}$ is the point at input, $\vect{y}$ is its farthest point in class, $\bar{\vect{x}}$ is the point at the output, and $\bar{\vect{z}}$ is its farthest point from 
the same class (it should not necessarily be the output of $\vect{y}$), then we calculate ${\norm{\bar{\vect{x}} - \bar{\vect{z}} }_2} - {\norm{\vect{x} - \vect{y}}_2}$ and ${\angle (\bar{\vect{x}},  \bar{\vect{z}}) } - {\angle(\vect{x}, \vect{y})}$. We do the same for the distances between different classes, comparing the shortest Euclidean and angular distances.}
\label{fig:imagenet_dist_dif_best}
\end{figure}

 \begin{figure}[bt]
\begin{center}
{
\subfigure[Inter-class Euclidean distance ratio]{\includegraphics[width=0.22\textwidth]{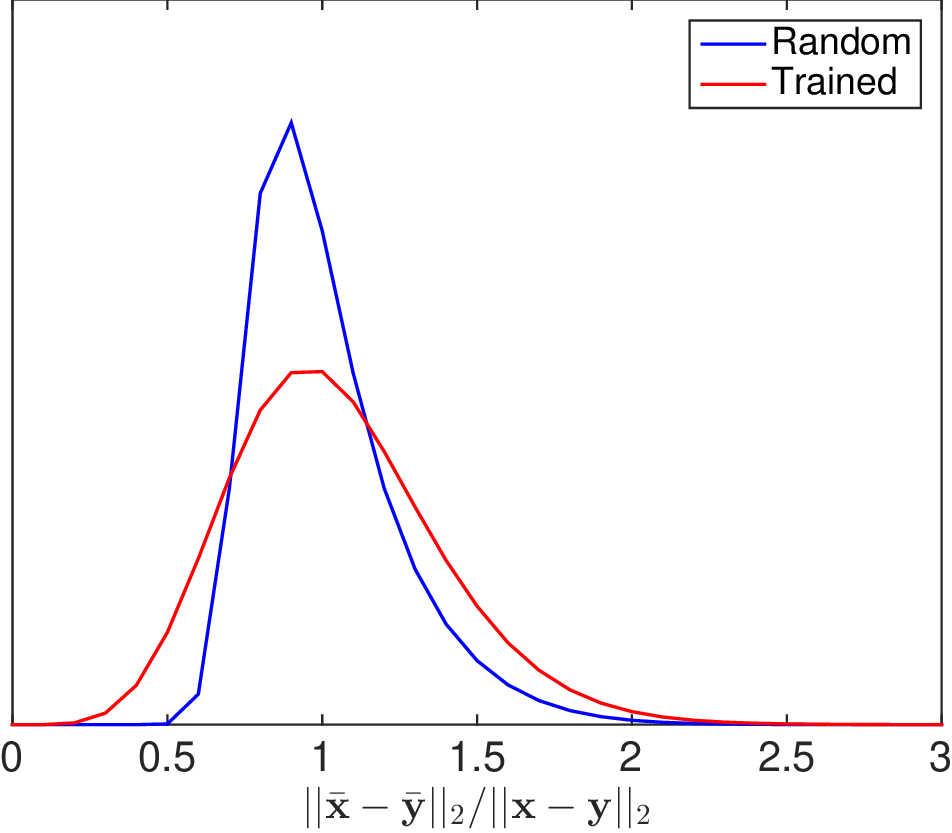}\label{fig:imagenet_euclid_all_diff}}\hfill
\subfigure[Intra-class Euclidean distance ratio]{\includegraphics[width=0.22\textwidth]{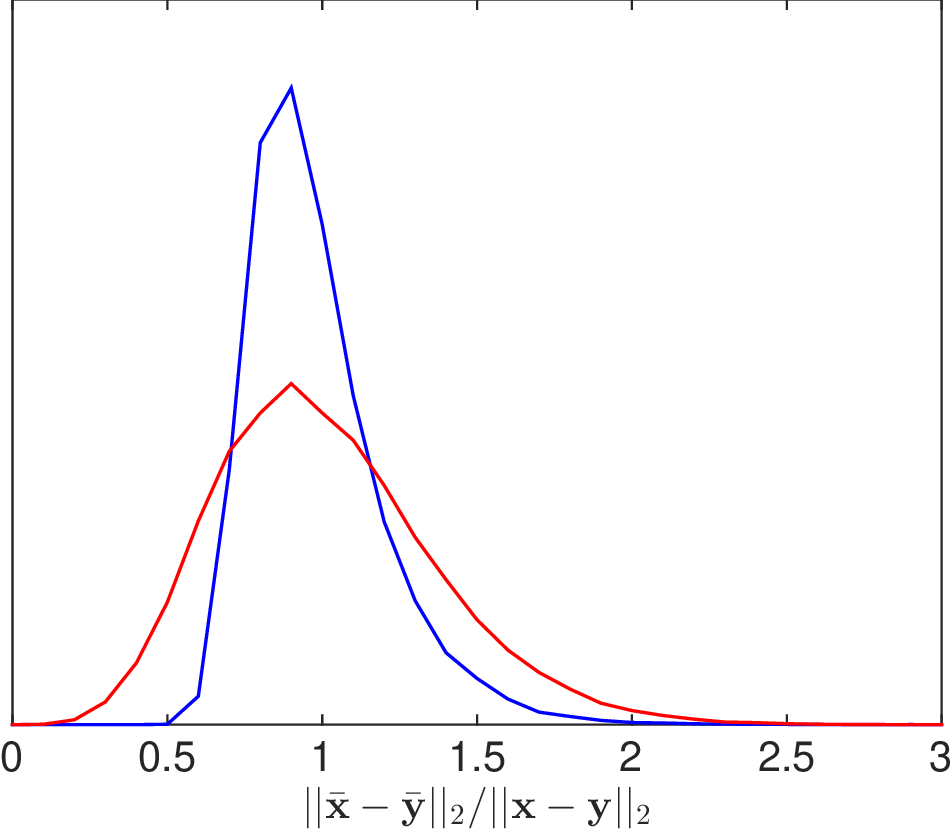}\label{fig:imagenet_euclid_all_same}}
\subfigure[Inter-class angular distance ratio]{\includegraphics[width=0.22\textwidth]{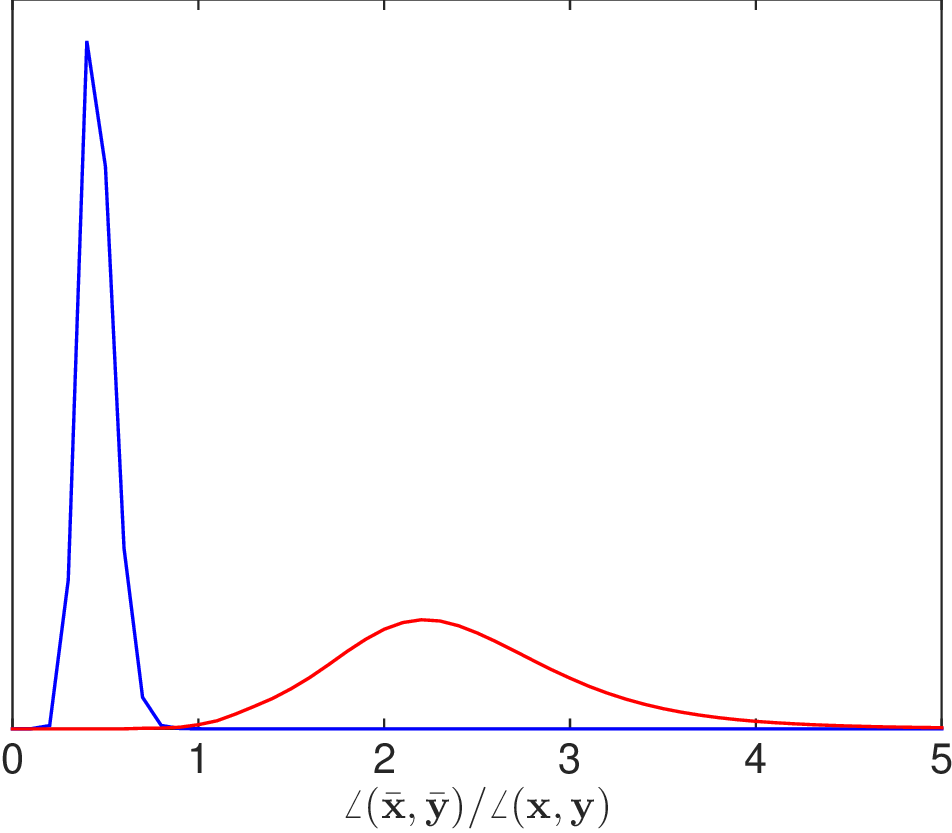}\label{fig:imagenet_angle_all_diff}}\hfill
\subfigure[Intra-class angular distance ratio]{\includegraphics[width=0.22\textwidth]{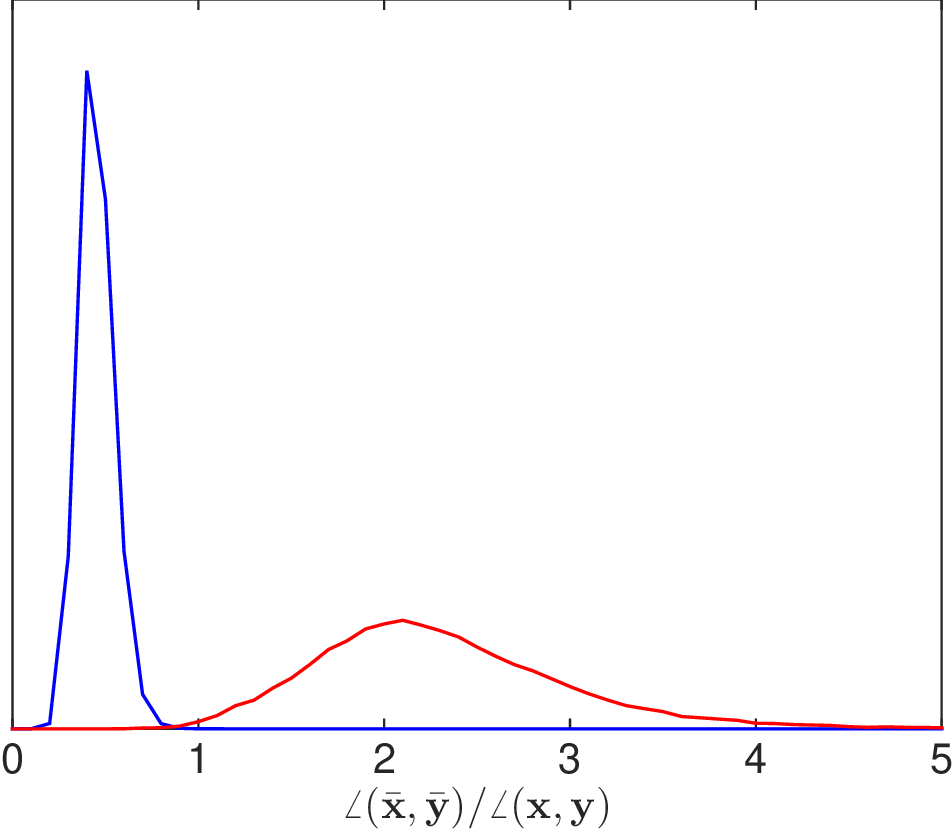}\label{fig:imagenet_angle_all_same}}}
\end{center}
\caption{Ratios of inter- (left) and intra- (right) class Euclidean (top) and angular (bottom) distances between \emph{randomly selected points} for ImageNet.
We calculate the Euclidean distances between randomly selected pairs of data points from different classes (left) and from the same class (right), both at the input of the DNN and at the output of the last convolutional layer.
Then we compute the ratio between the two, i.e., for all pairs of points $(\vect{x}, \vect{y})$ in the input and their corresponding points $(\bar{\vect{x}}, \bar{\vect{y}})$ at the output we calculate $\frac{\norm{\bar{\vect{x}} - \bar{\vect{y}} }_2}{\norm{\vect{x} - \vect{y}}_2}$ and $\frac{\angle (\bar{\vect{x}},  \bar{\vect{y}}) }{\angle(\vect{x}, \vect{y})}$.}
\label{fig:imagenet_dist_all}
\end{figure}

 \begin{figure}[bt]
\begin{center}
{
\hspace{0.1in}\subfigure[Inter-class Euclidean distance difference]{\includegraphics[width=0.22\textwidth]{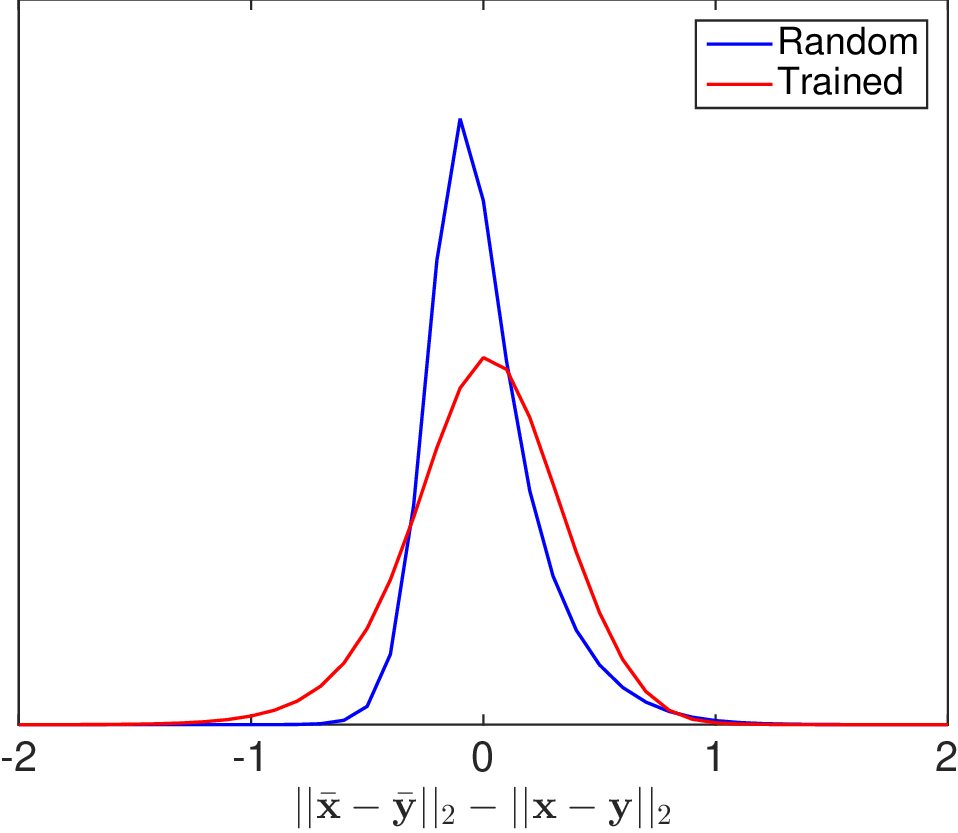}\label{fig:imagenet_euclid_dif_all_diff}}\hfill
\subfigure[Intra-class Euclidean distance difference]{\includegraphics[width=0.22\textwidth]{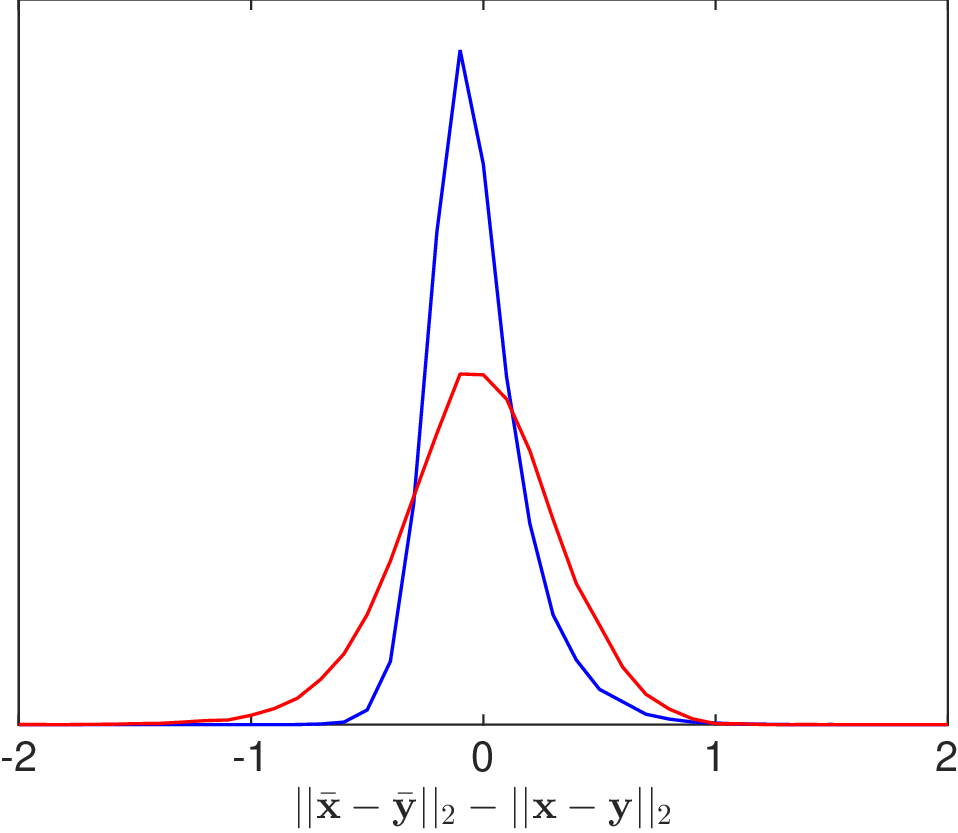}\label{fig:imagenet_euclid_dif_all_same}}
\hspace{0.08in}\subfigure[Inter-class angular distance difference]{\includegraphics[width=0.23\textwidth]{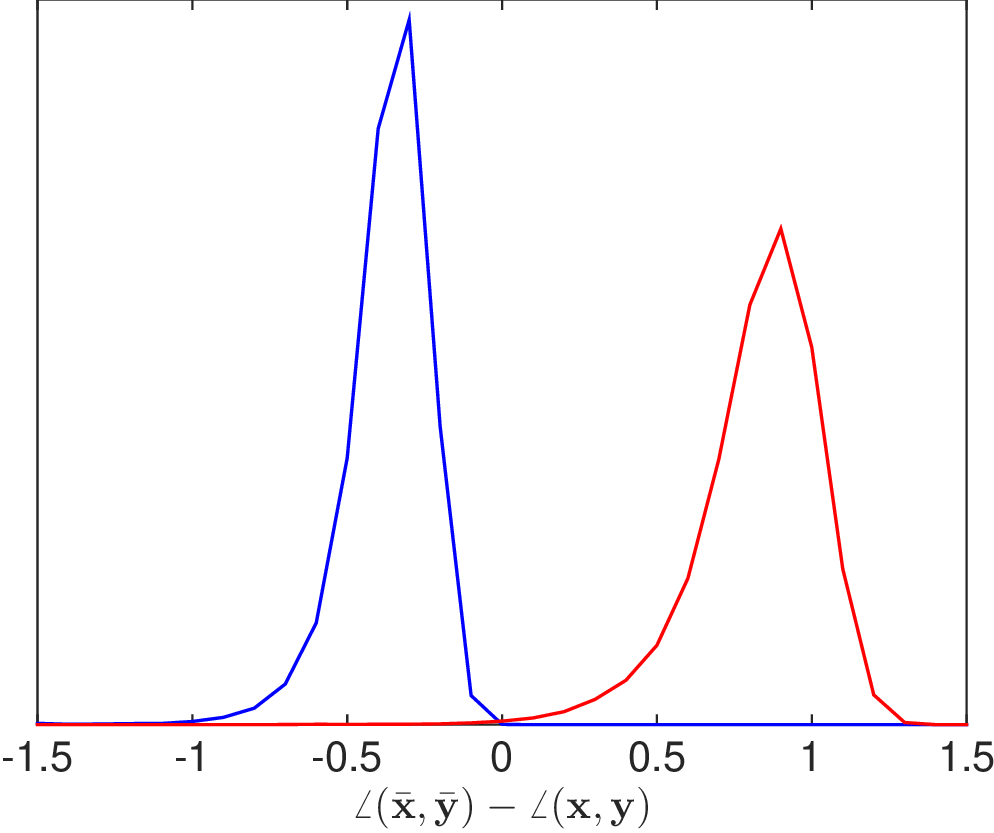}\label{fig:imagenet_angle_dif_all_diff}}\hfill
\subfigure[Intra-class angular distance difference]{\includegraphics[width=0.23\textwidth]{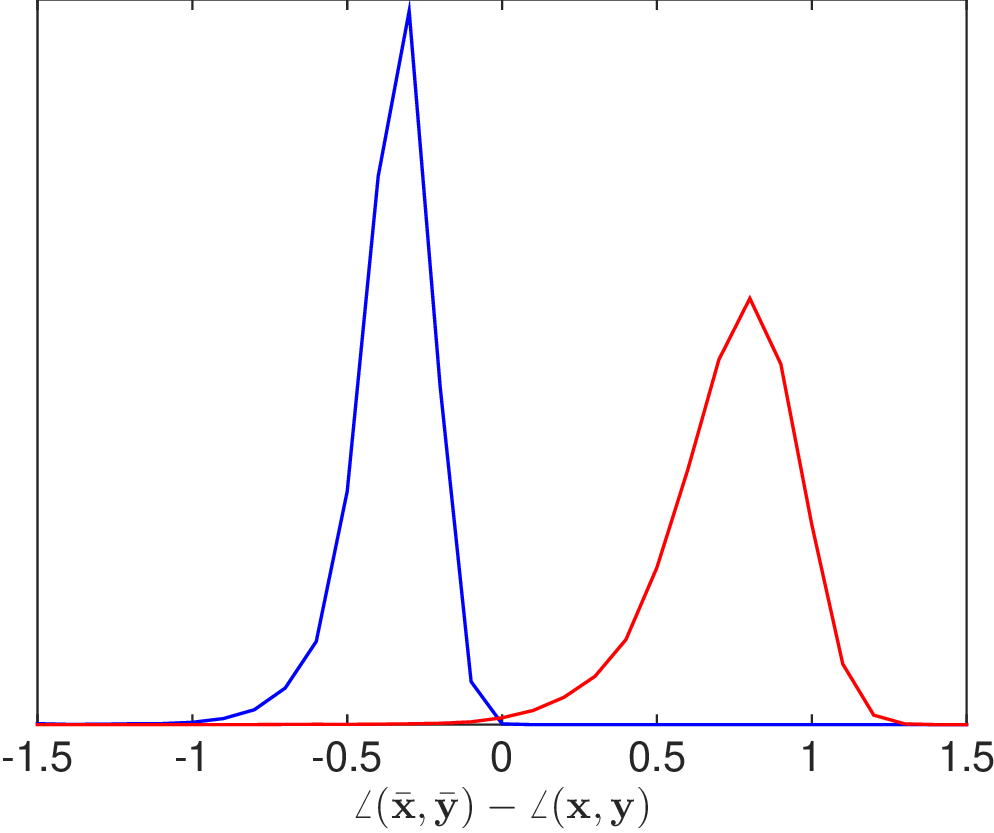}\label{fig:imagenet_angle_dif_all_same}}}
\end{center}
\caption{Differences of inter- (left) and intra- (right) class Euclidean (top) and angular (bottom) distances between \emph{randomly selected points} for ImageNet (a counterpart of Fig.~\ref{fig:imagenet_dist_all} with distance ratios replaced with differences).
We calculate the Euclidean distances between randomly selected pairs of data points from different classes (left) and from the same class (right), both at the input of the DNN and at the output of the last convolutional layer.
Then we compute the difference between the two, i.e., for all pairs of points $(\vect{x}, \vect{y})$ in the input and their corresponding points $(\bar{\vect{x}}, \bar{\vect{y}})$ at the output we calculate ${\norm{\bar{\vect{x}} - \bar{\vect{y}} }_2} - {\norm{\vect{x} - \vect{y}}_2}$ and ${\angle (\bar{\vect{x}},  \bar{\vect{y}})} - {\angle(\vect{x}, \vect{y})}$.}
\label{fig:imagenet_dist_dif_all}
\end{figure}

%
%
%

To validate this hypothesized behavior, we trained two DNN on the MNIST 
and CIFAR-10 
datasets, each containing $10$ classes.
The training of  the networks was done using the {\em matconvnet} toolbox \cite{Vedaldi15matconvnet}.
The MNIST and CIFAR-10 networks were trained with four and five layers, respectively, followed by a softmax operation. We used the default settings
 provided by the toolbox for each dataset, where with CIFAR-10 we also used horizontal mirroring and $64$ filters in the first two layers instead of $32$ (which is the default in the example provided with the package) to improve performance.
The trained DNN achieve $1\%$ and $18\%$ errors for MNIST and CIFAR-10 respectively. 

For each data point we calculate its Euclidean and angular distances to its farthest intra-class point and to its closest inter-class point.
We compute the ratio between the distances at the output of the last convolutional layer (the input of the fully connected layers) and the ones at the input. Let $\x$ be the point at the input, $\y$ be its farthest point from the same class, $\bar{\x}$ be the point at the output, and $\bar{\z}$ be its farthest point from the same class (it should not necessarily be the output of $\y$), then we calculate $\frac{\norm{\bar{\x} - \bar{\z} }_2}{\norm{\x - \y}_2}$ for Euclidean distances and $\angle(\bar\x, \bar\z ) /\angle(\x, \y )$ for the angles. We do the same for the distances between different classes, comparing the shortest ones.
Fig.~\ref{fig:cifar_dist_best} presents histograms of these distance ratios for CIFAR-10. In Fig.~\ref{fig:cifar_dist_dif_best} we present the histograms of the differences of the Euclidean and angular distances, i.e., 
${\norm{\bar{\x} - \bar{\z} }_2} - {\norm{\x - \y}_2}$  and $\angle(\bar\x, \bar\z )  - \angle(\x, \y ) $.
We also compare the behavior of all the inter and intra-class distances by  computing the above ratios for all pairs of points $(\x, \y)$ in the input  with respect to their corresponding points $(\bar\x, \bar\y)$ at the output.
These ratios are presented in Fig.~\ref{fig:cifar_dist_all}. We present also the differences ${\norm{\bar{\x} - \bar{\y} }_2} - {\norm{\x - \y}_2}$  and $\angle(\bar\x, \bar\y )  - \angle(\x, \y ) $
 in Fig.~\ref{fig:cifar_dist_dif_all}.
We present the results for three trained networks, in addition to the random one, denoted by Net1, Net2 and  Net3. Each of them corresponds to a different amount of training epochs, resulting with a different classification error. 

Considering the random DNN, note that all the histograms of the ratios are centered around $1$ and the ones of the differences around $0$, implying that the network preserves most of the distances as our theorems predict for a network with random weights.
For the trained networks, 
the histograms over all data point pairs (Figs.~\ref{fig:cifar_dist_all} and \ref{fig:cifar_dist_dif_all}) change only slightly due to training.
Also observe that the trained networks behave like their random counterparts in keeping the distance of a randomly picked pair of points.
However, they distort the distances between points on class boundaries
 ``better'' than the random network (Figs.~\ref{fig:cifar_dist_best} and \ref{fig:cifar_dist_dif_best}), in the sense that the farthest intra class distances are shrunk with a larger factor than the ones of the random network, and the closest inter class distances are set farther apart by the training. Notice that the shrinking of the distances within the class and enlargement of the distances between the classes improves as the training proceeds. 
This confirms our hypothesis that a goal of training is to treat the boundary points. 

A similar behavior can be observed for the angles. The closest angles are enlarged more in the trained network compared to the random one. 
However, enlarging the angles between classes also causes the enlargement of the angles within the classes. Notice though that these are enlarged less than the ones which are outside the class. Finally, observe that the enlargement of the angles, as we have seen in our theorems, causes a larger distortion in the Euclidean distances. Therefore, we may explain the enlargement of the distances in within the class as a means for shrinking the intra-class distances.

Similar behavior is observed for the MNIST dataset. However, the gaps between the random network and the trained network are smaller as the MNIST dataset contains  data which are  initially well separated. As we have argued above, for such manifolds the random network is already a good choice. 

We also compared the behavior of the validation data, of the ImageNet dataset, in the network provided by \cite{Simonyan15VeryDeep} and in  the same network but with random weights. The results are presented in Figs.~\ref{fig:imagenet_dist_best}, \ref{fig:imagenet_dist_dif_best},
\ref{fig:imagenet_dist_all} and \ref{fig:imagenet_dist_dif_all}. Behavior similar to 
 the one we observed in the case of CIFAR-10, is also manifested by the ImageNet network.

\section{Discussion and Conclusion}
\label{sec:conc}

We have shown that DNN with random Gaussian weights perform a stable embedding of the data, drawing a connection between the dimension of the features produced by the network that still keep the metric information of the original manifold, and the complexity of the data. 
The metric preservation property of the network provides a formal relationship between the complexity of the input data and the size of the required training set. Interestingly, follow-up studies \cite{Huang15Discriminative,Huang15Geometry} found that adding metric preservation constraints to the training of networks also leads to a theoretical relation between the complexity of the data and the number of training samples. Moreover, this constraint is shown to improve in practice the generalization error, i.e., improves the classification results when only a small number of training examples is available.

While preserving the structure of the initial metric is important, it is vital to have the ability to distort some of the distances in order to deform the data in a way that the Euclidean distances represent more faithfully the similarity we would like to have between points from the same class. We proved that such an ability is inherent to the DNN architecture: the Euclidean distances of the input data are distorted throughout the networks based on the angles between the data points.
Our results lead to the conclusion that DNN are universal classifiers for data based on the angles of the principal axis between the classes in the data. As these are not the angles we would like to work with in reality, the training of the DNN reveals the actual angles in the data. In fact, for some applications it is possible to use networks with  random weights at the first layers for separating the points with distinguishable angles, followed by trained weights at the deeper layers for separating the remaining points. This is practiced in the extreme learning machines (ELM) techniques \cite{Huang11Extreme} and our results provide a possible theoretical explanation for the success of this hybrid strategy.


Our work implies that it is possible to view DNN as a stagewise metric learning process, suggesting that it might be possible to replace the currently used layers with other metric learning algorithms, opening a new venue for semi-supervised DNN. 
This also stands in line with the recent literature on convolutional kernel methods (see \cite{Mairal2014Convolutional,Lu14How}).

In addition, we observed that a potential main goal of the training of the network is to treat the class boundary points, while keeping the other distances approximately the same. This may lead to a new active learning strategy for deep learning \cite{Elhamifar13Convex}.


\noindent
{\bf Acknowledgments-} Work partially supported by NSF, ONR, NGA, NSSEFF, and ARO.  A.B. is supported by ERC StG 335491 (RAPID).  The authors thank the reviewers
of the manuscript for their suggestions which greatly improved the paper.

\appendices
\section{Proof of Theorem~\ref{thm:DNN_euclidean_layeri} }
\label{sec:proof_euclid}

Before we turn to prove Theorem~\ref{thm:DNN_euclidean_layeri}, we present {two}
propositions  that will aid us in its proof.
The first is the Gaussian concentration bound that appears in \cite[Equation 1.6]{Talagrand91Probability}.

\begin{prop}
\label{prop:Gauss_concentration_bound}
Let $\vect{g}$ be an i.i.d. Gaussian random vector with zero mean and unit variance, and $\eta$ be a Lipschitz-continuous function with a Lipschitz constant $c_\eta$. Then for every $\alpha > 0$, with probability exceeding $1 - 2\exp(-\alpha^2/2c_\eta)$,
\begin{eqnarray}
\abs{ \eta(\vect{g}) - \mathbb{E}[\eta(\vect{g})]} < \alpha.
\end{eqnarray}
\end{prop}

\begin{prop}
\label{prop:max_bound}
Let $\m \in \RR{n}$ be a random vector with zero-mean i.i.d. Gaussian distributed entries with variance $\frac{1}{m}$, and $K \subset \mathbb{B}_1^n$ be a set with a Gaussian mean width $w(K)$. Then, with probability exceeding $1 - 2\exp(- \omega(K)^2/4)$,
\begin{eqnarray}
\label{eq:sup_max_bound}
\sup_{\x,\y \in K}\left(\rho(\m^T\x) - \rho(\m^T\y)\right)^2 <4 \frac{\omega(K)^2}{m}.
\end{eqnarray}
\end{prop}
{\it Proof:} 
First, notice that from the properties of the ReLU $\rho$, it holds that 
\begin{eqnarray}
\label{eq:rhogTx_rhogTy_ineq}
\abs{\rho(\m^T\x) - \rho(\m^T\y)} \le \abs{\m^T\x - \m^T\y}  =  \abs{\m^T(\x - \y)}.
\end{eqnarray}
Let $\g = \sqrt{m}\cdot \m$ be a scaled version of $\m$ such that each entry in $\g$ has a unit variance (as $\m$ has a variance $\frac{1}{m}$).
From the Gaussian mean width charachteristics (see Proposition~2.1 in \cite{Plan13Robust}), we have $\mathbb{E} \sup_{\x, \y \in K}\abs{\g^T(\x - \y)} = w(K)$. Therefore, combining the Gaussian concentration bound in Proposition~\ref{prop:Gauss_concentration_bound} together with the fact that $\sup_{\x, \y \in K}\abs{\g^T (\x - \y) }$ is Lipschitz-continuous with a constant $c_{\eta} = 2$ (since $K \subset \mathbb{B}_1^n$), we have
\begin{eqnarray}
\label{eq:gTxy_omega_ineq_alpha}
\abs{ \sup_{\x, \y \in K}\g^T(\x - \y) - \omega(K)} < \alpha,
\end{eqnarray}
with probability exceeding $(1 - 2\exp(- \alpha^2/4))$.
Clearly, \eqref{eq:gTxy_omega_ineq_alpha} implies
\begin{eqnarray}
\label{eq:gTxy_ineq_omega_alpha}
\sup_{\x, \y \in K}{\g^T(\x - \y)} \le 
 2\omega(K),
\end{eqnarray}
where we set $\alpha = \omega(K)$.
Combining \eqref{eq:gTxy_ineq_omega_alpha} and \eqref{eq:rhogTx_rhogTy_ineq} with the fact that $$\sup_{\x,\y \in K}\left(\rho(\g^T\x) - \rho(\g^T\y)\right)^2  = (\sup_{\x,\y \in K}\left(\rho(\g^T\x) - \rho(\g^T\y)\right))^2$$  
leads to 
\begin{eqnarray}
\sup_{\x,\y \in K}\left(\rho(\g^T\x) - \rho(\g^T\y)\right)^2 < 4\omega(K)^2.
\end{eqnarray}
Dividing both sides by $m$
completes the proof.
 \hfill $\Box$

{\it Proof of Theorem~\ref{thm:DNN_euclidean_layeri}:} 
Our proof of Theorem~\ref{thm:DNN_euclidean_layeri} consists of three keys steps. In the first one, we show that the bound in \eqref{eq:DNN_euclidean_bound_layeri} holds with  high probability for any two points $\x, \y \in K$. In the second, we pick an $\epsilon$-cover for $K$ and show that the same holds for each pair in the cover. The last generalizes the bound for any point in $K$.

{\bf Bound for a pair $\x,\y \in K$:}
Denoting by $\m_i$ the $i$-th column of $\M$, we rewrite 
\begin{eqnarray}
\norm{\rho(\matr{M}\vect{x}) - \rho(\matr{M}\vect{y})}_2^2  = \sum_{i}^m \left(\rho(\m_i^T\x) - \rho(\m_i^T\y)\right)^2.
\end{eqnarray}
Notice that since all the $\m_i$ have the same distribution, the random variables $\left(\rho(\m_i^T\x) - \rho(\m_i^T\y)\right)^2$ are also equally-distributed. Therefore, our strategy would be to calculate the expectation of these random variables and then to show, using Bernstein's inequality, that the mean of these random variables does not deviate much from their expectation.

We start by calculating their expectation
\begin{eqnarray}
\label{eq:euc_dist_expecation_stg1}
&& \hspace{-0.3in} \mathbb{E}\left[\left(\rho(\m_i^T\x) - \rho(\m_i^T\y)\right)^2\right] \\ \nonumber && \hspace{-0.1in}  = 
\mathbb{E}\left(\rho(\m_i^T\x) \right)^2  + \mathbb{E}\left(\rho(\m_i^T\y)\right)^2 - 2\mathbb{E} \rho(\m_i^T\x)\rho(\m_i^T\y).
\end{eqnarray}
For calculating the first term at the right hand side (rhs) note that $\m_i^T\x$ is a random Gaussian vector with variance $\norm{\x}_2^2/m$. Therefore, from the symmetry of the Gaussian distribution we have that 
\begin{eqnarray}
\mathbb{E}\left(\rho(\m_i^T\x) \right)^2 = \frac{1}{2}\mathbb{E}\left(\m_i^T\x \right)^2 =\frac{1}{2m} \norm{\x}_2^2. 
\end{eqnarray}
In the same way, $\mathbb{E}\left(\rho(\m_i^T\y) \right)^2 = \frac{1}{2m} \norm{\y}_2^2$. For calculating the third term at the rhs of \eqref{eq:euc_dist_expecation_stg1}, notice that $\rho(\m_i^T\x)\rho(\m_i^T\y)$ is non-zero if both the inner product between $\m_i$ and $\x$ and the inner product between $\m_i$ and $\y$ are positive. Therefore, the smaller the angle between $\x$ and $\y$, the higher the probability that both of them will have a positive inner product with $\m_i$. Using the fact that a Gaussian vector is uniformly distributed on the sphere, we can calculate the expectation of $\rho(\m_i^T\x)\rho(\m_i^T\y)$ by the following integral, which is dependent on the angle between $\x$ and $\y$:
\begin{eqnarray}
\label{eq:rhmiTx_rhomiTy_calc_angle}
&&\hspace{-0.25in} \mathbb{E}[\rho(\m_i^T\x)\rho(\m_i^T\y)]=  \\ \nonumber && ~~
\frac{\norm{\x}_2\norm{\y}_2}{m\pi} \int_0^{\pi - \angle(\x,\y) } \sin(\theta)\sin(\theta + \angle(\x,\y)) d\theta
\\ \nonumber && ~~ = \frac{\norm{\x}\norm{\y} }{m\pi}\left(\sin(\angle (\x, \y))  - 
 \cos(\angle (\x, \y) ) \angle (\x, \y)-\pi\right).
\end{eqnarray}
Having the expectation of all the terms in \eqref{eq:euc_dist_expecation_stg1} calculated, we define the following random variable, which is the difference between $\left(\rho(\m_i^T\x) - \rho(\m_i^T\y)\right)^2$ and its expectation,
\begin{eqnarray}
\label{eq:z_i}
&&   \hspace{-0.3in} z_i \triangleq \left(\rho(\m_i^T\x) - \rho(\m_i^T\y)\right)^2 -\frac{1}{m}
\bigg(\frac{1}{2}\norm{\x -\y }_2^2   \\  && \nonumber  \hspace{-0.3in}
 + \frac{\norm{\x}_2\norm{\y}_2}{\pi}\Big( \sin(\angle (\x, \y))  - 
 \cos(\angle (\x, \y) ) \angle (\x, \y)
 \Big).
\end{eqnarray}
Clearly, the random variable $z_i$ is zero-mean. To finish the first step of the proof, it remains to show that the sum $\sum_{i=1}^m z_i$ does not deviate much from zero (its mean). First, note that 
\begin{eqnarray}
\label{eq:z_i_sum_traingle_inequality}
P\left( \abs{\sum_{i=1}^m {z_i} }> t\right) \le P\left( \sum_{i=1}^m \abs{z_i} > t\right),
\end{eqnarray}
and therefore it is enough to bound the term on the rhs of \eqref{eq:z_i_sum_traingle_inequality}.
By Bernstein's inequality, we have 
\begin{eqnarray}
\label{eq:Bernstein}
P\left( \sum_{i=1}^m \abs{z_i} > t\right) \le \exp\left(- \frac{t^2/2}{\sum_{i=1}^m \mathbb{E}z_i^2 + Mt/3} \right),
\end{eqnarray}
where $M$ is an upper bound on $\abs{z_i}$. 
To calculate $\mathbb{E}z_i^2$, one needs to calculate the fourth moments $\mathbb{E}(\rho(\m_i^T\x))^4$ and $\mathbb{E}(\rho(\m_i^T\y))^4$, which is easy to compute by using the symmetry of Gaussian vectors,
and the correlations $\mathbb{E}\rho(\m_i^T\x)^3\rho(\m_i^T\y)$, $\mathbb{E}\rho(\m_i^T\x)\rho(\m_i^T\y)^3$ and $\mathbb{E}\rho(\m_i^T\x)^2\rho(\m_i^T\y)^2$. For calculating the later, we use as before the fact that a Gaussian vector is uniformly distributed on the sphere and calculate an integral on an interval which is dependent on the angle between $\x$ and $\y$. For example,
\begin{eqnarray}
\label{eq:ERhox3Rhoy}
&& \hspace{-0.4in} \mathbb{E}\left[\rho(\m_i^T\x)^3\rho(\m_i^T\y)\right] = \frac{4\norm{\x}^3_2\norm{\y}_2}{m^2\pi}\cdot \\ \nonumber && ~~~~~~~
\int_0^{\pi - \angle(\x,\y) } \sin^3(\theta)\sin(\theta + \angle(\x,\y)) d\theta,
\end{eqnarray}
where $\theta$ is the angle between $\m_i$ and $\x$. We have a similar formula for the other terms. By simple arithmetics and using the fact that $K \in \mathbb{B}_1^n$, we have that $\mathbb{E}z_i^2 \le 2.1/m^2$.

The type of formula in \eqref{eq:ERhox3Rhoy}, which is similar to the one in \eqref{eq:rhmiTx_rhomiTy_calc_angle}, provides an insight into the role of training. As random layers `integrate uniformly' on the interval $[0, \pi - \angle(\x,\y)]$, learning picks the angle $\theta$ that maximizes/minimizes the inner product based on whether $\x$ and $\y$ belong to the same class or to distinct classes.

Using Proposition~\ref{prop:max_bound}, the fact that $K \in \mathbb{B}_1^n$ and the behavior of $\dist(\x,\y)$ (see Fig.~\ref{fig:cos_sin_term_behav}) together with the triangle inequality imply that $M < \frac{4\omega(K)^2+3}{m}$ with probability exceeding $(1-2\exp(-\omega(K)^2/4))$. Plugging this bound with the one we computed for $Ez_i^2$ into \eqref{eq:Bernstein} leads to
\begin{eqnarray}
 \hspace{-0.2in} P\left( \sum_{i=1}^m \abs{z_i} > \frac{\delta}{2} \right) &  \hspace{-0.1in} \le &  \hspace{-0.1in}  \exp\left(-\frac{m\delta^2 /8}{2.1^2 + 4\delta\omega(K)^2/3+\delta} \right) \\ \nonumber &&  \hspace{-0.1in}
 + 2\exp(-\omega(K)^2/4), 
\end{eqnarray}
where we included the probability of Proposition~\ref{prop:max_bound} in the above bound. Since by the assumption of the theorem $m = O(\delta^{-4}\omega(K)^4)$, we can write
\begin{eqnarray}
\label{eq:z_i_t_xy_inequality}
P\left( \sum_{i=1}^m \abs{z_i} > \frac{\delta}{2}\right) \le C\exp\left(- \frac{m\delta^2}{4w(K)^2} \right).
\end{eqnarray}

{\bf Bound for all $\x,\y \in N_\epsilon(K)$:} Let $N_\epsilon(K)$ be an $\epsilon$ cover for $K$.
By using a union bound we have that  for every pair in $N_\epsilon(K)$,
\begin{eqnarray}
\label{eq:z_i_t_net_inequality_stg1}
P\left( \sum_{i=1}^m \abs{z_i} > \frac{\delta}{2}\right) \le C\abs{N_{\epsilon}(K)}^2\exp\left(- \frac{m\delta^2}{4w(K)^2} \right).
\end{eqnarray}
By Sudakov's inequality we have $\log\abs{N_\epsilon(K)} \le c\epsilon^{-2}w(K)^2$.  Plugging this inequality into \eqref{eq:z_i_t_net_inequality_stg1} leads to
\begin{eqnarray}
\label{eq:z_i_t_net_inequality_stg2}
P\left( \sum_{i=1}^m \abs{z_i} > \frac{\delta}{2}\right) \le C\exp\left(- \frac{m\delta^2}{4w(K)^2 }+c\epsilon^{-2}w(K)^2 \right).
\end{eqnarray}
Setting $\epsilon \ge \frac{1}{50}\delta$, we have by the assumption $m \ge C \delta^{-4}\omega(K)^2$ that the term in the exponent at the rhs of \eqref{eq:z_i_t_net_inequality_stg2} is negative and therefore the probability decays exponentially as  $m$ increases.

{\bf Bound for all $\x,\y \in K$}: Let us rewrite $\x,\y \in K$ as  $\x_0 + \x_{\epsilon}$ and $\y_0 + \y_{\epsilon}$, with $\x_0, \y_0 \in N_{\epsilon}(K)$ and $\x_{\epsilon}, \y_{\epsilon}  \in (K - K) \cap \mathbb{B}_\epsilon^n$, where $K - K  = \left\{ \x - \y : \x, \y \in K \right\}$. We get to the desired result by  setting $\epsilon < \frac{1}{40}\delta$ and using the triangle inequality combined with Proposition~\ref{prop:max_bound} to control $\rho(\M\x_\epsilon)$ and $\rho(\M\y_{\epsilon})$, the fact that $w(K-K) \le 2w(K)$ and the Taylor expansions of the $\cos(\cdot), \sin(\cdot)$ and $\cos^{-1}(\cdot)$ functions to control the terms related to $\psi$ (see   \eqref{cor:DNN_euclidean_layeri_liph}).
 \hfill $\Box$

\section{Proof of Theorem~\ref{thm:DNN_angle_layeri}}
\label{sec:proof_angle}

{\it Proof:} 
Instead of proving Theorem~\ref{thm:DNN_angle_layeri} directly, we  deduce it from Theorem~\ref{thm:DNN_euclidean_layeri}. 
First we notice that \eqref{eq:DNN_euclidean_bound_layeri} is equivalent to 
\begin{eqnarray}
\label{eq:DNN_euclidean_bound_layeri_ang_equiv}
&& \hspace{-0.3in} \Bigg\vert \frac{1}{2}\norm{\rho(\matr{M}\vect{x})}_2^2 + \frac{1}{2}\norm{\rho(\matr{M}\vect{y})}_2^2 - \frac{1}{4}\norm{\x}_2^2 - \frac{1}{4}\norm{\y }_2^2 
\\  && \hspace{-0.3in} \nonumber
  - \Big( \rho(\matr{M}\vect{x})^T \rho(\matr{M}\vect{y}) - \frac{\norm{\x}_2\norm{\y}_2}{2}\cos(\angle (\x, \y) ) -   \\  \nonumber && \hspace{-0.3in}
\frac{\norm{\x}_2\norm{\y}_2}{2\pi} \big( \sin(\angle (\x, \y))  - 
 \cos(\angle (\x, \y) ) \angle (\x, \y)
 \big) \Big)
\Bigg\vert  \le \frac{\delta}{2}.
\end{eqnarray}
As $E\norm{\rho(\matr{M}\vect{x})}_2^2 = \frac{1}{2}\norm{\x}_2^2$, we  also have that with high probability (like the one in Theorem~\ref{thm:DNN_euclidean_layeri}),
\begin{eqnarray}
\label{eq:rho_M_euclidean_bound_signle}
\abs{\norm{\rho(\matr{M}\vect{x})}_2^2 - \frac{1}{2}\norm{\x}_2^2} \le \delta, & \forall \x \in K.
\end{eqnarray}
(The proof is very similar to the one of Theorem~\ref{thm:DNN_euclidean_layeri}).
Applying the reverse triangle inequality to \eqref{eq:DNN_euclidean_bound_layeri_ang_equiv} and then using  \eqref{eq:rho_M_euclidean_bound_signle}, followed by dividing both sides by 
 $\frac{\norm{\x}_2\norm{\y}_2}{2}$,  lead to
\begin{eqnarray}
\label{eq:DNN_ang_bound_layeri_div_norm_xy}
&& \hspace{-0.3in} \Bigg\vert  \Big( \frac{2\rho(\matr{M}\vect{x})^T \rho(\matr{M}\vect{y})}{\norm{\x}_2\norm{\y}_2} - \cos(\angle (\x, \y) ) -   \\  \nonumber && \hspace{-0.3in}
\frac{1}{\pi} \big( \sin(\angle (\x, \y))  - 
 \cos(\angle (\x, \y) ) \angle (\x, \y)
 \big) \Big)
\Bigg\vert  \le \frac{3\delta}{\norm{\x}_2\norm{\y}_2}.
\end{eqnarray}
Using the reverse triangle inequality with \eqref{eq:DNN_ang_bound_layeri_div_norm_xy} leads to
\begin{eqnarray}
\label{eq:DNN_ang_bound_layeri_div_norm_xy_stg2}
&& \hspace{-0.3in} \Bigg\vert  \Big( \frac{\rho(\matr{M}\vect{x})^T \rho(\matr{M}\vect{y})}{\norm{\rho(\M\x)}_2\norm{\rho(\M\y)}_2} - \cos(\angle (\x, \y) ) -   \\  \nonumber && \hspace{-0.3in}
\frac{1}{\pi} \big( \sin(\angle (\x, \y))  - 
 \cos(\angle (\x, \y) ) \angle (\x, \y)
 \big) \Big)
\Bigg\vert  \le \frac{3\delta}{\norm{\x}_2\norm{\y}_2} \\ && \nonumber + \abs{\frac{2\rho(\matr{M}\vect{x})^T \rho(\matr{M}\vect{y})}{\norm{\x}_2\norm{\y}_2} - \frac{\rho(\matr{M}\vect{x})^T \rho(\matr{M}\vect{y})}{\norm{\rho(\M\x)}_2\norm{\rho(\M\y)}_2}}.
\end{eqnarray}
To complete the proof it remains to bound the rhs of \eqref{eq:DNN_ang_bound_layeri_div_norm_xy_stg2}. For the second term in it, we have
\begin{eqnarray}
\label{eq:rhoxrhoy_div_xy_bound}
&&  \hspace{-0.2in} \abs{\frac{2\rho(\matr{M}\vect{x})^T \rho(\matr{M}\vect{y})}{\norm{\x}_2\norm{\y}_2} - \frac{\rho(\matr{M}\vect{x})^T \rho(\matr{M}\vect{y})}{\norm{\rho(\M\x)}_2\norm{\rho(\M\y)}_2}} = \\ && \nonumber 
 \hspace{-0.2in} 
\frac{2\rho(\matr{M}\vect{x})^T \rho(\matr{M}\vect{y})}{\norm{\x}_2\norm{\y}_2} \abs{  1 -   \frac{\norm{\x}_2\norm{\y}_2}{2\norm{\rho(\M\x)}_2\norm{\rho(\M\y)}_2} }.
\end{eqnarray}
Because $K \subset \mathbb{B}_1^n \setminus \beta \mathbb{B}_2^n$, it follows from \eqref{eq:rho_M_euclidean_bound_signle} that 
\begin{eqnarray}
\label{eq:norm_rhoMx_upperbound}
\norm{\rho(\M\x)}_2^2 \ge \frac{1}{2}\beta^2 - \delta.
\end{eqnarray} 
Dividing by $\left(\norm{\rho(\M\x)}_2 + \frac{1}{\sqrt{2}}\norm{\x}_2\right)\norm{\rho(\M\x)}_2$
both sides of \eqref{eq:rho_M_euclidean_bound_signle}  and then using \eqref{eq:norm_rhoMx_upperbound} and the fact that $\norm{\x}_2 \ge \beta$, provide
\begin{eqnarray}
\label{eq:rhoMx_div_x_bound_1p_delta}
&& \hspace{-0.3in}\abs{  1 -   \frac{\norm{\x}_2}{\sqrt{2}\norm{\rho(\M\x)}_2} }  
\\ && \nonumber \hspace{-0.1in}
\le \frac{\delta}{\left(\norm{\rho(\M\x)}_2 + \frac{1}{\sqrt{2}}\norm{\x}_2\right)\norm{\rho(\M\x)}_2}
\\ && \nonumber \hspace{-0.1in} \le \frac{\delta}{\left(\sqrt{\frac{1}{2}\beta^2 - \delta} + \frac{1}{\sqrt{2}}\beta\right)\sqrt{\frac{1}{2}\beta^2 - \delta}}
\le \frac{\delta}{\beta^2 - 2\delta},
 \forall \x \in K,
\end{eqnarray}
where the last inequality is due to simple arithmetics.
Using the triangle inequality and then the fact that the inequality in \eqref{eq:rhoMx_div_x_bound_1p_delta} holds $\forall \x \in K$, we have
\begin{eqnarray}
\label{eq:1_min_xy_div_rhoMx_rhoy_ineq_bound}
&& \hspace{-0.3in}\abs{  1 -   \frac{\norm{\x}_2\norm{\y}_2}{2\norm{\rho(\M\x)}_2\norm{\rho(\M\y)}_2}} 
\le \abs{  1 -   \frac{\norm{\y}_2}{\sqrt{2}\norm{\rho(\M\y)}_2}}  \\ && \nonumber
~~~~~~~~~~~~~~+\abs{  1 -   \frac{\norm{\x}_2}{\sqrt{2}\norm{\rho(\M\x)}_2}}\cdot\abs{ \frac{\norm{\y}_2}{\sqrt{2}\norm{\rho(\M\y)}_2}}  \\ && \nonumber
~~\le \frac{\delta}{\beta^2 - 2\delta} + \frac{\delta}{\beta^2 - 2\delta}(1+ \frac{\delta}{\beta^2 - 2\delta}) \le \frac{3\delta}{\beta^2 - 2\delta} .
\end{eqnarray}
Combining \eqref{eq:DNN_ang_bound_layeri_div_norm_xy} with the facts that $\cos(\angle (\x, \y) ) +  
\frac{1}{\pi} \big( \sin(\angle (\x, \y))  - 
 \cos(\angle (\x, \y) ) \angle (\x, \y)$ is bounded by $1$ (see Fig.~\ref{fig:cos_sin_term_behav}) and $K \subset \mathbb{B}_1^n \setminus \beta \mathbb{B}_2^n$ lead to
\begin{eqnarray}
\label{eq:rhoMxrhoMy_xy_ineq_1p_delta_div_beta}
\frac{2\rho(\matr{M}\vect{x})^T \rho(\matr{M}\vect{y})}{\norm{\x}_2\norm{\y}_2} \le \left(1+ \frac{3\delta}{\beta^2} \right).
\end{eqnarray}
Plugging \eqref{eq:rhoMxrhoMy_xy_ineq_1p_delta_div_beta} and \eqref{eq:1_min_xy_div_rhoMx_rhoy_ineq_bound} into \eqref{eq:rhoxrhoy_div_xy_bound} and then the outcome into   \eqref{eq:DNN_ang_bound_layeri_div_norm_xy_stg2} lead to the bound $\left(1+ \frac{3\delta}{\beta^2} \right)\frac{3\delta}{\beta^2 - 2\delta} + \frac{3\delta}{\beta^2} \le \frac{15\delta}{\beta^2 - 2\delta}$
 \hfill $\Box$

\bibliographystyle{IEEEtran}
\bibliography{deep_learn}

\end{document}